\documentclass[10pt,twocolumn,letterpaper]{article}

\usepackage{cvpr}
\usepackage{graphicx}
\usepackage{amsmath}
\usepackage{amssymb}
\usepackage{booktabs}
\usepackage{multirow}
\usepackage{colortbl}
\usepackage{xcolor}
\usepackage{microtype}
\usepackage{pifont}
\usepackage{enumitem}
\setlist[itemize]{leftmargin=1.2em,itemsep=1pt,topsep=2pt,parsep=0pt}
\setlist[enumerate]{leftmargin=1.4em,itemsep=1pt,topsep=2pt,parsep=0pt}

\usepackage{xcolor}

\definecolor{revCmjX}{RGB}{31,78,144}
\definecolor{revQoTu}{RGB}{46,130,65}
\definecolor{rev8Gov}{RGB}{137,71,145}

\definecolor{cvprblue}{rgb}{0.21,0.49,0.74}
\definecolor{rowgray}{gray}{0.95}
\definecolor{bestblue}{RGB}{228,240,252}
\definecolor{mygreen}{RGB}{17,128,41}
\definecolor{myred}{RGB}{185,0,27}
\newcommand{\cmark}{\textcolor{mygreen}{\ding{51}}}
\newcommand{\xmark}{\textcolor{myred}{\ding{55}}}

\usepackage[pagebackref,breaklinks,colorlinks,allcolors=cvprblue]{hyperref}
\usepackage[capitalize]{cleveref}

\AtBeginDocument{%
  \setlength{\abovedisplayskip}{4pt plus 1pt minus 1pt}%
  \setlength{\belowdisplayskip}{4pt plus 1pt minus 1pt}%
  \setlength{\abovedisplayshortskip}{2pt plus 1pt minus 1pt}%
  \setlength{\belowdisplayshortskip}{2pt plus 1pt minus 1pt}%
}
\def\confName{CVPR}
\def\confYear{2027}

\title{\emph{Sfumato} Supervision: Learning Instance Segmentation with Noisy Labels}

\author{
  \textbf{Moshe Kimhi\textsuperscript{1}},
  \textbf{Ehud Rivlin\textsuperscript{1}},
  \textbf{Chaim Baskin\textsuperscript{2}},
\\
  \textsuperscript{1}Technion,
  \textsuperscript{2}Ben-Gurion University
  \\
\href{https://mkimhi.github.io/Sfumato}{\texttt{https://mkimhi.github.io/Sfumato}}
}

\begin{document}
\maketitle

\begin{abstract}
Instance segmentation requires precise object masks, yet both human annotators and promptable models can produce unreliable contours. A high prediction loss near a contour does not necessarily indicate an annotation error: valid edges can also be uncertain. Inspired by Leonardo da Vinci's \emph{sfumato}, which softens transitions between a subject and its surroundings, we introduce \textbf{\texttt{CB-LNL} (Copula-Boundary Learning with Noisy Labels)}. Rather than judging each contour pixel alone, CB-LNL combines foreground evidence inside the mask with background evidence outside it to guide local supervision and annotation repair. We ground this approach in an audit of paired COCO and LVIS annotations, and introduce benchmarks covering human annotations, weak-prompt SAM2 masks, and controlled corruptions in natural and driving scenes. Controlled comparisons show improved training on procedural contour noise and fewer harmful edits than direct teacher-mask replacement. Paired-seed and frozen-prediction evaluations reveal how these benefits depend on the annotation source, supporting separate assessment of category and contour reliability.
\end{abstract}

\begin{figure}[t]
    \centering
    \includegraphics[width=\linewidth]{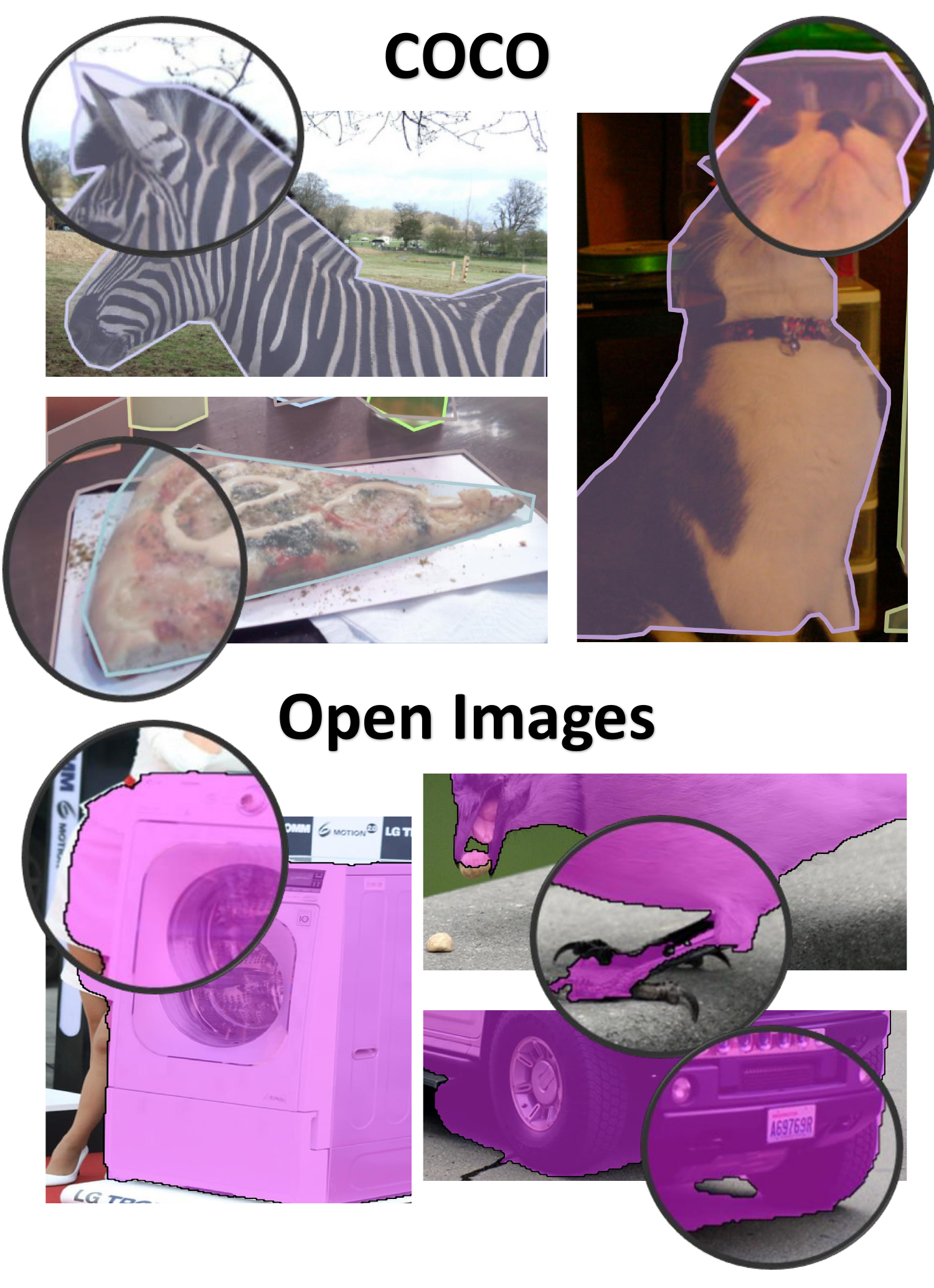}
    \caption{\textbf{Annotation errors in practice.} COCO~\cite{lin2014microsoft} and Open Images~\cite{OpenImages} examples show coarse contours, masks extending into the background, category errors, and missing or merged instances. These errors affect different parts of the supervision and need not make the entire annotation unusable.}
    \label{fig:teaser_errors}
\end{figure}

\begin{figure*}[t]
    \centering
    \includegraphics[width=0.96\textwidth]{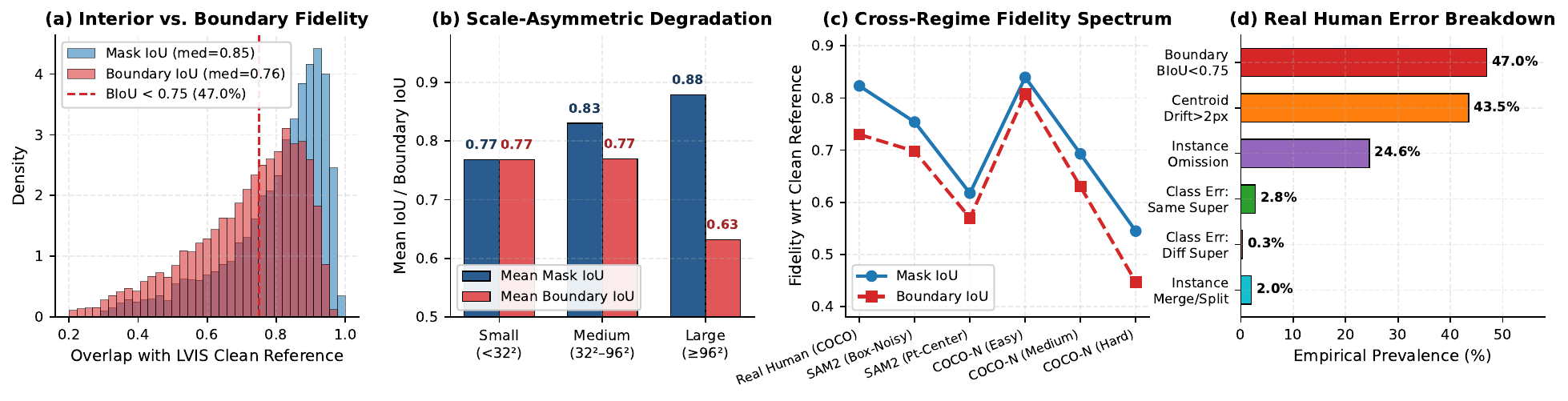}
    \caption{\textbf{Human annotation audit on 24,809 shared COCO--LVIS images.} \textbf{(a)} Same-class pairs can have high Mask IoU but lower Boundary IoU; $47.0\%$ have $\text{BIoU}<0.75$. \textbf{(b)} Small objects have lower Mask IoU, while large objects have lower Boundary IoU. \textbf{(c)} Annotation agreement across human masks, SAM2 masks, and procedural corruptions. \textbf{(d)} Frequencies of the audited error types. Matching rules and evaluation populations are in \cref{sec:appendix_audit}.}
    \label{fig:audit}
\end{figure*}

\begin{figure*}[t]
    \centering
    \includegraphics[width=0.95\textwidth]{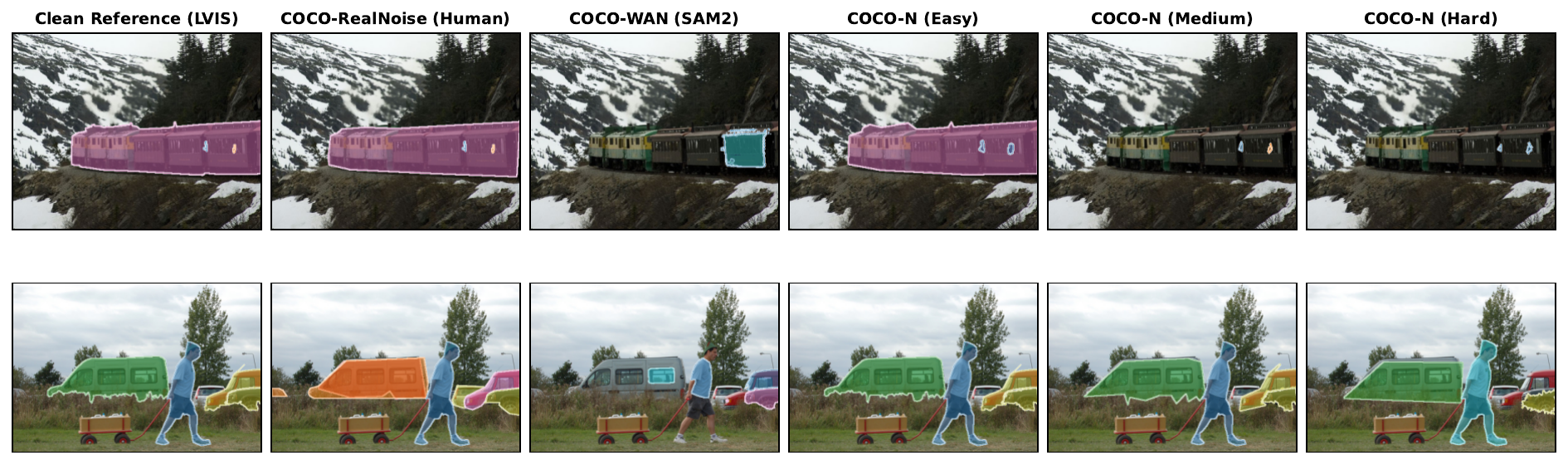}
    \caption{\textbf{The three annotation sources on the same images.} Columns show reference masks, original COCO annotations (COCO-RealNoise), point-prompt SAM2 masks (COCO-WAN), and Easy, Medium, and Hard procedural corruptions (COCO-N). Additional driving examples, isolated corruptions, and repair cases appear in \cref{fig:supp_viper,fig:supp_operators,fig:supp_cases}.}
    \label{fig:visual_taxonomy}
\end{figure*}

\section{Introduction}
\label{sec:intro}

Leonardo da Vinci used \emph{sfumato} to soften the transition between a subject and its surroundings~\cite{davinci1651trattato,gombrich1995story}. Instance segmentation faces a related challenge: a binary mask must place a precise contour even where the image is ambiguous. In practice, human annotators simplify outlines, misplace contours, confuse categories, and miss objects (\cref{fig:teaser_errors}). Promptable models such as SAM2 can reduce annotation effort, but weak prompts introduce errors of their own, including masks that cover only part of an object or extend into the background~\cite{kirillov2023segany,ravi2024sam2}.

Learning with Noisy Labels (LNL) has been studied extensively for image classification~\cite{Song2020LearningFN,han2018coteaching,li2020dividemix,zhang2018gce,wang2019sce,wei2022learning}. In that setting, each image has a single category label. A segmentation annotation can instead contain a correct category and an incorrect contour, or vice versa. Existing noisy-segmentation studies examine important parts of this problem~\cite{yang2020lncis,Liu_2022,yao2023learning}, but a joint evaluation of human masks, model-generated masks, and controlled contour errors is still needed (\cref{tab:prior_comparison}). Evaluation also matters: agreement with the original annotations can reward the same shortcuts that a model learns during training. We call the resulting reversal between original-annotation and reference-annotation rankings the \textbf{Noisy Validation Illusion} (\cref{sec:exp_suite1}).

\begin{figure*}[t]
\centering
\includegraphics[width=0.94\textwidth]{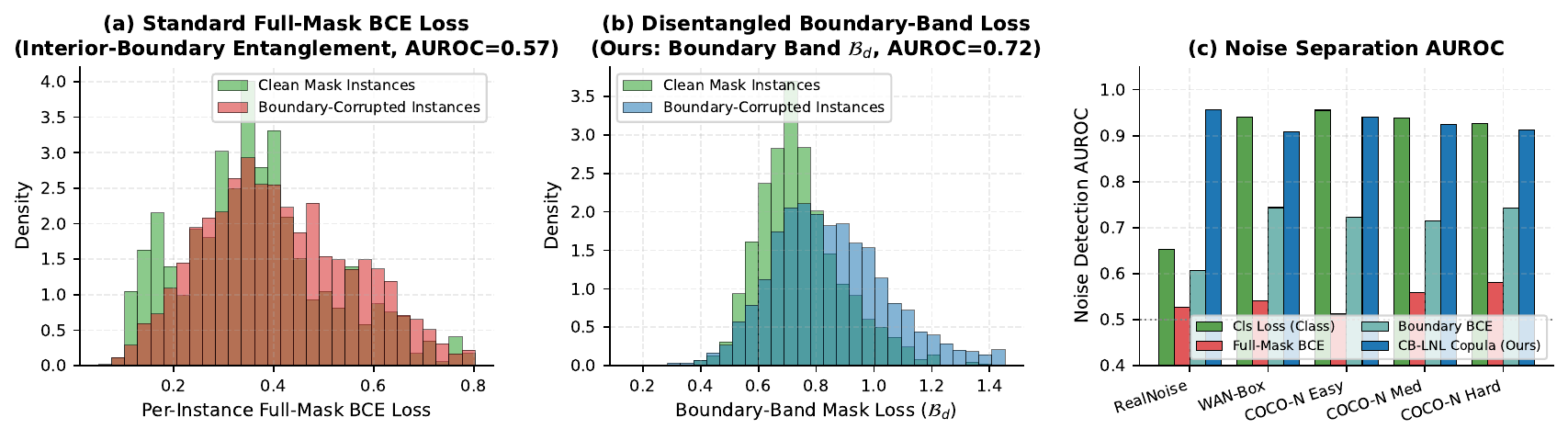}
\caption{\textbf{Category and contour errors produce different loss signals.} \textbf{(a)} Full-mask BCE distributions overlap for clean and contour-corrupted instances (plotted AUROC $0.57$). \textbf{(b)} Boundary-band BCE separates them more clearly (plotted AUROC $0.72$), but valid uncertain edges still overlap with errors. \textbf{(c)} Noise-detection AUROC across settings under this diagnostic protocol. These results motivate separate category and contour scores; the controlled comparison on identical frozen predictions is reported in \cref{tab:suite4_copula}.}
\label{fig:loss_entanglement}
\end{figure*}

Our loss analysis suggests why a single reliability score is insufficient (\cref{fig:loss_entanglement}). Classification loss can reveal category errors, but averaging mask loss over the full object and surrounding background can hide a small contour error. Measuring loss only near the contour makes such errors easier to detect. Yet a valid edge can also have high loss when the model is uncertain. The missing information is spatial context: does the predicted foreground continue across the annotated contour, or does it give way to background?

This observation motivates \textbf{\emph{Sfumato} Supervision}. Our method, \textbf{\texttt{CB-LNL} (Copula-Boundary Learning with Noisy Labels)}, examines predictions on both sides of an annotated contour. Foreground inside and background outside support the existing edge; agreement on foreground or background across both sides suggests a misplaced contour. Our \textbf{Tri-Coupled Fr\'echet Copula} combines interior, exterior, and category--spatial evidence to guide local supervision and annotation repair (\cref{sec:method}). We make three contributions:
\begin{itemize}
    \item \textbf{An audit and benchmark suite.} We compare COCO masks with LVIS re-annotations and study SAM2 masks under four prompt settings. These observations guide benchmarks spanning human annotations, weak-prompt masks, and procedural corruptions (\cref{sec:audit,sec:benchmark}).
    \item \textbf{Boundary-aware noisy-label learning.} We separate category and contour reliability, then use directional context to refine supervision near uncertain contours (\cref{sec:method}).
    \item \textbf{Controlled evaluation of detection, training, and repair.} We compare scoring rules on frozen predictions, repeat training across paired seeds, and compare local repair with direct teacher-mask replacement (\cref{sec:experiments}).
\end{itemize}

\begin{table}[t]
\centering
\caption{\textbf{Comparison with Prior Noisy-Annotation Settings in Dense Prediction.} ``Audit'' denotes calibration and evaluation against verified human re-annotations; ``Contour'' denotes geometrically correlated border noise; ``Clean'' denotes automated dataset cleaning.}
\label{tab:prior_comparison}
\resizebox{\columnwidth}{!}{%
\begin{tabular}{llccccccc}
\toprule
\textbf{Study} & \textbf{Task} & \textbf{Audit} & \textbf{Contour} & \textbf{Scale} & \textbf{Drift} & \textbf{Omit} & \textbf{WAN} & \textbf{Clean} \\
\midrule
Yang \etal~\cite{yang2020lncis} & Inst. & \xmark & \xmark & \xmark & \xmark & \xmark & \xmark & \xmark \\
Liu \etal~\cite{Liu_2022} & Sem. & \xmark & \cmark & \cmark & \xmark & \xmark & \xmark & \xmark \\
Yao \etal~\cite{yao2023learning} & Sem. & \xmark & \cmark & \cmark & \xmark & \xmark & \xmark & \xmark \\
Cityscapes-C~\cite{cordts2016cityscapes} & Sem/Inst & \xmark & \cmark & \cmark & \xmark & \xmark & \xmark & \xmark \\
SAM SA-1B~\cite{kirillov2023segany} & Prompt & \xmark & \cmark & \xmark & \xmark & \xmark & \cmark & \xmark \\
\rowcolor{bestblue}\textbf{Ours} & \textbf{Inst.} & \cmark & \cmark & \cmark & \cmark & \cmark & \cmark & \cmark \\
\bottomrule
\end{tabular}%
}
\end{table}

\section{Related Work}
\label{sec:related}

\noindent\textbf{Noisy-label benchmarks.}
Classification benchmarks study synthetic category flips and human label errors, including CIFAR-N~\cite{wei2022learning} and Clothing1M~\cite{xiao2015learning}. Detection studies also consider inaccurate boxes and missing objects~\cite{mao2021noisy,ryoo2023universal,liu2022robust,tkachenko2023objectlab}. For segmentation, prior work examines category errors and spatially corrupted masks~\cite{yang2020lncis,Liu_2022,yao2023learning,Lu2014LearningFW,Li2023,nordstrm2022image}. Our benchmark brings together paired human annotations, weak-prompt masks, and controlled corruptions so that these sources can be evaluated within one study.

\noindent\textbf{Weak and model-generated annotations.}
Cityscapes-Coarse~\cite{cordts2016cityscapes}, Open Images~\cite{OpenImages}, and SA-1B~\cite{kirillov2023segany} illustrate different ways to reduce annotation effort. We study a complementary question: how does prompt quality affect the masks used to train an instance segmentation model? COCO-WAN addresses this question using SAM2~\cite{ravi2024sam2} with tight boxes, perturbed boxes, center points, and random points.

\noindent\textbf{Robust learning and contextual refinement.}
GCE~\cite{zhang2018gce} and SCE~\cite{wang2019sce} modify classification losses, while Co-teaching~\cite{han2018coteaching} and DivideMix~\cite{li2020dividemix} use loss-based sample selection. Our experiments test adaptations of these ideas to instance segmentation. S4MC~\cite{kimhi2023semisupervised} uses neighboring predictions to refine pseudo-label confidence in semi-supervised semantic segmentation. CB-LNL also uses local context, but distinguishes predictions inside and outside an annotated mask and combines them to assess contour reliability.

\section{Annotation Noise in Practice}
\label{sec:audit}
We study two annotation sources: human masks in MS-COCO~\cite{lin2014microsoft} and masks generated by SAM2~\cite{ravi2024sam2}.

\subsection{Human Annotations: COCO and LVIS}
\label{sec:human_audit}
COCO and LVIS share images but provide independently collected object annotations~\cite{lin2014microsoft,gupta2019lvis}. We map their categories using WordNet synsets and compare annotations on \textbf{24,809 shared images}. The audit contains \textbf{59,840 object matches}; contour analysis uses \textbf{57,962 same-class pairs}. LVIS masks serve as the higher-quality reference, rather than an assumption that every reference contour is exact. Matching thresholds and eligible categories are specified in \cref{sec:appendix_audit}. Four patterns emerge (\cref{fig:audit,tab:audit_fidelity}).

\noindent\textbf{1. Contour disagreement is common.}
Among same-class matched pairs, mean Mask IoU is $0.823$, whereas mean Boundary IoU (BIoU, dilation ratio $d=0.02$~\cite{Cheng_2021}) is $0.730$. In this population, $47.0\%$ of masks have $\text{BIoU}<0.75$, and $18.9\%$ have $\text{BIoU}<0.60$. Substantial area overlap can therefore coexist with a poorly aligned contour.

\noindent\textbf{2. COCO polygons use fewer vertices and have displaced centroids.}
COCO masks use a median of $42.4\%$ as many vertices as their LVIS references (interquartile range $27.3\%$--$64.0\%$). Their centroids differ by a median of $1.65$ pixels, and $43.5\%$ differ by more than $2$ pixels. These measurements motivate separate tests of polygon simplification and spatial displacement; centroid displacement alone does not identify a rigid translation.

\noindent\textbf{3. Agreement depends on object size and metric.}
Small objects ($<32^2$ pixels) have lower mean Mask IoU than large objects ($\ge96^2$ pixels): $0.769$ versus $0.879$. The ordering reverses for Boundary IoU: $0.768$ for small objects and $0.632$ for large objects. Area-based and contour-based metrics thus reveal different weaknesses (\cref{fig:audit}b).

\noindent\textbf{4. Category disagreement and missing instances are structured.}
Among pairs with $\text{IoU}\ge0.70$, $3.1\%$ have different category labels; $88.9\%$ of these disagreements are within the same supercategory. Within LVIS-verified categories, COCO omits $24.6\%$ of reference instances. Merged instances account for $1.6\%$ of masks.

\begin{table*}[t]
\centering
\caption{\textbf{Effects of annotation source across backbones.} Each backbone has separate Mask AP and Boundary AP columns. Models use the fixed reference validation annotations; parentheses for COCO-RealNoise instead show evaluation against original COCO annotations. VIPER-N does not include ConvNeXt-Tiny. Complete metrics appear in \cref{tab:suite1_archs_full}.}
\label{tab:suite1_archs}
\begingroup
\setlength{\tabcolsep}{4pt}
\resizebox{0.8\textwidth}{!}{%
\begin{tabular}{llrrrrrr}
\toprule
 & & \multicolumn{2}{c}{\textbf{ResNet-50}} & \multicolumn{2}{c}{\textbf{ConvNeXt-Tiny}} & \multicolumn{2}{c}{\textbf{Swin-Tiny}} \\
\cmidrule(lr){3-4}\cmidrule(lr){5-6}\cmidrule(lr){7-8}
\textbf{Dataset} & \textbf{Supervision} & Mask AP & Bnd. AP & Mask AP & Bnd. AP & Mask AP & Bnd. AP \\
\midrule
COCO-RealNoise & Reference & 30.89 (28.31) & 17.56 (16.41) & 31.26 (28.76) & 17.85 (16.58) & 31.34 (28.71) & 17.86 (16.57) \\
 & Original COCO & 30.45 (29.83) & 17.05 (16.82) & 30.69 (30.22) & 17.00 (16.90) & 30.94 (30.59) & 17.16 (17.07) \\
\midrule
COCO-WAN & Tight box & 30.87 & 18.21 & 31.71 & 18.51 & 31.69 & 18.76 \\
 & Perturbed box & 28.63 & 16.57 & 28.98 & 16.98 & 29.17 & 17.05 \\
 & Center point & 26.21 & 15.04 & 26.76 & 15.56 & 26.47 & 15.32 \\
 & Random point & 25.00 & 14.23 & 25.43 & 14.57 & 25.56 & 14.79 \\
\midrule
COCO-N & Easy & 29.09 & 16.29 & 30.10 & 16.91 & 29.55 & 16.70 \\
 & Medium & 26.70 & 14.34 & 26.57 & 14.25 & 27.51 & 14.97 \\
 & Hard & 24.09 & 10.48 & 24.67 & 10.97 & 24.51 & 10.75 \\
\midrule
VIPER-N & Clean & 24.19 & 15.93 & -- & -- & 24.75 & 16.34 \\
 & Easy & 21.86 & 14.12 & -- & -- & 22.42 & 14.56 \\
 & Medium & 19.65 & 11.82 & -- & -- & 20.14 & 12.24 \\
 & Hard & 17.56 & 9.76 & -- & -- & 18.02 & 10.12 \\
\bottomrule
\end{tabular}%
}
\endgroup
\end{table*}

\subsection{Weak-Prompt SAM2 Annotations}
\label{sec:wan_audit}
SAM2 produces different errors depending on its prompt (\cref{fig:audit}c and \cref{tab:audit_fidelity}). With tight reference boxes, its mean Boundary IoU is slightly higher than that of COCO polygons ($0.741$ versus $0.730$), although its mean Mask IoU is lower ($0.796$ versus $0.823$). With perturbed boxes, $25.6\%$ of small-object masks exceed $1.5$ times the reference area. Single-point prompts often select only part of a large object, such as a shirt instead of a person (\cref{fig:visual_taxonomy}). Mean Boundary IoU for large objects falls to $0.430$ with center points and $0.410$ with random points. These patterns motivate separate box- and point-prompt settings in COCO-WAN.

\section{Benchmark Design}
\label{sec:benchmark}

\begin{table}
\centering
\caption{\textbf{Effects of individual corruptions with ResNet-50-FPN.} Each row isolates one operator, except the final rows, which apply scale-normalized mixtures. Displacement uses $\mu=4$ pixels and vertex jitter uses $\sigma=2.5$ pixels. Category-flip and omission rates are $20\%$. Erosion/dilation uses odd-sized kernels; ``50/50'' randomly chooses either direction. Slash-separated results follow the order in the row label.}
\label{tab:suite2_ablations}
\resizebox{\columnwidth}{!}{%
\begin{tabular}{lccccccc}
\toprule
\textbf{Isolated Operator} & $\textbf{AP}^{\textbf{mask}}\uparrow$ & $\textbf{AP}^{\textbf{bnd}}\uparrow$ & $\textbf{AP}^{\textbf{box}}\uparrow$ & $\textbf{AP}_{\textbf{S}}\uparrow$ & $\textbf{AP}_{\textbf{M}}\uparrow$ & $\textbf{AP}_{\textbf{L}}\uparrow$ & $\textbf{ECE}\downarrow$ \\
\midrule
Clean Reference & \textbf{30.10} & \textbf{17.10} & \textbf{31.86} & \textbf{16.38} & \textbf{35.18} & \textbf{42.18} & 19.00 \\
\midrule
Polygon Simplification & 27.07 & 14.03 & 30.15 & 11.45 & 33.51 & 41.66 & 18.95 \\
Rigid Spatial Drift & \textbf{25.14} & \textbf{11.43} & 28.83 & \textbf{8.24} & 31.72 & 40.56 & 18.31 \\
Vertex Jitter & 29.67 & 16.36 & 27.91 & 15.26 & 34.62 & 41.13 & 18.66 \\
\midrule
Morphological Erosion & \textbf{20.85} & \textbf{8.59} & \textbf{25.17} & \textbf{5.73} & \textbf{24.94} & 38.47 & 20.52 \\
Morphological Dilation & \textbf{22.88} & \textbf{8.63} & 28.34 & 6.24 & 26.47 & 39.31 & 20.05 \\
Unbiased Scale (50/50) & \textbf{28.23} & \textbf{15.29} & 29.75 & 13.18 & 34.68 & 41.31 & 18.51 \\
\midrule
Uniform  & 28.80  & 16.67 & 30.08& 15.56 & 35.97 & 40.34  & 13.82 \\
Super-Cat Flip &  28.94 & 16.85 &  30.50 &  15.56 &  33.54 &  41.26 & \textbf{13.23} \\
Instance Omission  & 28.76  & 15.76  & 29.78  & 15.11 & 33.15  & 41.35  & 17.49 \\
Instance Merging &  27.96 & 15.61 &  28.70 & 14.81 & 34.32 &  39.61 &  17.98 \\
\midrule
Scale-Norm Easy & 29.23  & 15.70  & 30.60  & 15.33  & 35.11  & 41.33  & 16.25 \\
Scale-Norm Med & 26.21  & 12.93  & 27.81  &  15.29  &  29.87  &  37.80  & 17.36 \\
Scale-Norm Hard &  \textbf{21.62} &  \textbf{8.16} & 25.19 &  13.31 &  25.43 &  \textbf{30.71} &15.19 \\

\bottomrule
\end{tabular}%
}
\end{table}

\begin{table*}[t]
\centering
\caption{\textbf{Robust training under noisy annotations (single seed).} Results are Mask AP / Boundary AP. WAN-Box and WAN-Point denote perturbed boxes and center points, respectively. Three-seed results are reported separately in \cref{tab:suite3_seeds}.}
\label{tab:suite3_lnl}
\begingroup
\setlength{\tabcolsep}{4pt}
\resizebox{0.98\textwidth}{!}{%
\begin{tabular}{lrrrrrr}
\toprule
\textbf{Method} & \textbf{RealNoise} & \textbf{WAN-Box} & \textbf{WAN-Point} & \textbf{Easy} & \textbf{Medium} & \textbf{Hard} \\
 & \multicolumn{6}{c}{Mask AP / Boundary AP (single seed)} \\
\midrule
Standard & 30.45 / 17.05 & 28.63 / 16.57 & 26.21 / 15.04 & 29.09 / 16.29 & 26.70 / 14.34 & 24.09 / 10.48 \\
SCE~\cite{wang2019sce} & 29.52 / 15.92 & 27.17 / 15.10 & 25.30 / 14.41 & 28.91 / 16.14 & 26.22 / 13.73 & 23.16 / 10.09 \\
GCE~\cite{zhang2018gce} & 30.01 / 16.40 & 28.53 / 16.29 & 25.94 / 15.26 & 29.68 / 16.86 & 27.21 / 14.34 & 23.07 / 10.13 \\
DivideMix-Cls~\cite{li2020dividemix} & 27.06 / 15.09 & 25.10 / 14.17 & 22.75 / 13.38 & 27.24 / 15.15 & 25.25 / 13.02 & 21.92 / 9.86 \\
DivideMix-Mask & 28.32 / 15.33 & 27.44 / 15.53 & 25.78 / 14.99 & 29.40 / 16.57 & 26.01 / 13.92 & 23.39 / 10.39 \\
Spatial self-correction & 29.87 / 16.53 & 27.69 / 15.68 & 25.32 / 14.28 & 28.91 / 15.90 & 26.82 / 14.47 & 23.51 / 10.71 \\
Boundary-LNL & 30.54 / 17.19 & 28.50 / 16.71 & 25.97 / 15.16 & 29.68 / 16.86 & 28.30 / 15.48 & 24.97 / 11.45 \\
\rowcolor{bestblue}CB-LNL & 30.87 / 17.22 & 28.14 / 16.48 & 26.07 / 15.78 & 30.01 / 17.19 & 28.45 / 15.33 & 25.62 / 11.25 \\
\bottomrule
\end{tabular}%
}

\endgroup
\end{table*}

\begin{table}[t]
\centering
\caption{\textbf{Three-seed training results.} Mean $\pm$ standard deviation across three paired seeds, with splits, corruptions, and initializations matched within each seed.}
\label{tab:suite3_seeds}
\begingroup
\small
\setlength{\tabcolsep}{3pt}
\begin{tabular*}{0.85\columnwidth}{@{\extracolsep{\fill}}lrr@{}}
\toprule
\textbf{Method} & \textbf{Mask AP} & \textbf{Boundary AP} \\
\midrule
\multicolumn{3}{@{}c}{\textbf{COCO-RealNoise}} \\
Standard & $30.19\pm0.26$ & $16.87\pm0.17$ \\
Boundary-LNL & $\mathbf{30.85\pm0.34}$ & $\mathbf{17.44\pm0.22}$ \\
\rowcolor{bestblue}CB-LNL & $30.54\pm0.35$ & $17.25\pm0.08$ \\
\midrule
\multicolumn{3}{@{}c}{\textbf{COCO-N Medium}} \\
Standard & $27.48\pm0.83$ & $14.60\pm0.28$ \\
Boundary-LNL & $28.12\pm0.16$ & $15.31\pm0.17$ \\
\rowcolor{bestblue}CB-LNL & $\mathbf{28.57\pm0.27}$ & $\mathbf{15.37\pm0.04}$ \\
\bottomrule
\end{tabular*}
\endgroup
\end{table}

\begin{table*}
\centering
\caption{\textbf{Error scoring and training ablations.} \textbf{Top:} All scoring rules use identical frozen predictions. FPR is measured on valid uncertain edges at $80\%$ true-positive rate; paired FPR values are COCO-N Medium / COCO-RealNoise. \textbf{Bottom:} Separate training runs on COCO-N Medium vary the refinement components (\cref{tab:suite4_full_supp}).}
\label{tab:suite4_copula}
\resizebox{0.88\textwidth}{!}{%
\begin{tabular}{lccccc}
\toprule
\textbf{Frozen Scoring Rule (Same Checkpoint)} & \multicolumn{2}{c}{\textbf{COCO-N Med} ($N=2507$)} & \multicolumn{2}{c}{\textbf{RealNoise} ($N=1123$)} & \textbf{Edge FPR$\downarrow$} \\
\cmidrule(lr){2-3} \cmidrule(lr){4-5}
 & \textbf{AUROC}$\uparrow$ & \textbf{PR-AUC}$\uparrow$ & \textbf{AUROC}$\uparrow$ & \textbf{PR-AUC}$\uparrow$ & \textbf{Med / Real} \\
\midrule
Whole-Mask BCE ($\ell_{\text{mask}}^{\text{full}}$) & 0.7143 & 0.7625 & 0.6412 & 0.6745 & 36.7\% / 71.2\% \\
Boundary-Band BCE ($\ell_{\text{bnd}}$ on $\mathcal{B}_1$) & \textbf{0.7915} & \textbf{0.8518} & 0.6290 & 0.6621 & 28.5\% / 69.3\% \\
Linear Contrast ($p_{\text{in}} - p_{\text{out}}$) & 0.6856 & 0.7287 & 0.6334 & 0.6616 & 40.0\% / \textbf{55.8\%} \\
\rowcolor{bestblue}\textbf{\texttt{CB-LNL} Fr\'echet Copula (Ours)} & 0.7880 & 0.8511 & \textbf{0.6469} & \textbf{0.6776} & \textbf{23.3\%} / 61.9\% \\
\midrule
\textbf{Training Ablation (COCO-N Med)} & \textbf{Copula $\lambda$} & $\textbf{AP}^{\textbf{mask}}\uparrow$ & $\textbf{AP}^{\textbf{bnd}}\uparrow$ & $\textbf{AP}^{\textbf{box}}\uparrow$ & $\textbf{AP}_{\textbf{L}}\uparrow$ \\
\midrule
Standard / \texttt{Boundary-LNL} & -- & 26.70 / 28.30 & 14.34 / \textbf{15.48} & 28.46 / 29.74 & 39.11 / \textbf{41.61} \\
Isotropic Band ($\mathcal{B}_d$) / Class--Spatial ($C_\lambda$) & $\hat{\lambda}=0.53$ & 27.66 / 28.21 & 14.85 / 15.39 & 29.36 / \textbf{30.02} & 40.61 / 41.00 \\
Interior ($p_{\text{in}}$) / Exterior  ($1-p_{\text{out}}$) & $\hat{\lambda}=0.53$ & 28.10 / 27.46 & 15.44 / 14.60 & 29.79 / 29.27 & 41.31 / 39.33 \\
Fixed ~\cite{kimhi2023semisupervised}  & $\hat{\lambda}=0$ & 27.96  & 15.02  & \textbf{30.04} & 40.35  \\
Fixed   & $\hat{\lambda}=1$ & 27.71  &  15.14 &  29.96  & 41.15  \\
 Hard Flip & $\hat{\lambda}=0.8$ &  26.89 &  14.88 &  28.70 & 38.83 \\

\rowcolor{bestblue}\textbf{\texttt{CB-LNL} (Full Tri-Coupled)} & $\hat{\lambda}=0.53$ & \textbf{28.45} & 15.33 & 29.87 & 41.22 \\
\bottomrule
\end{tabular}%
}
\end{table*}

\begin{table*}[t]
\centering
\caption{\textbf{Annotation repair and full-scale training.} \textbf{Top:} Single-stage repair and teacher-mask replacement use identical frozen auditor predictions and a matched $30\%$ selection budget. Multi-stage results are marked separately. For COCO human masks, Boundary IoU uses LVIS-verified pairs; full-split reference agreement is in parentheses. Downstream AP uses the fixed reference validation set. Frozen-scoring AUROCs in teacher-replacement rows follow \cref{tab:suite4_copula}; the other auditing AUROCs follow \cref{tab:supp_cross_audit}. \textbf{Bottom:} Full-scale training on COCO-N Medium and Hard.}
\label{tab:suite5_cleaning}
\label{tab:suite6_full}
\resizebox{\textwidth}{!}{%
\begin{tabular}{llcccccccc}
\toprule
\textbf{Dataset} & \textbf{Supervision / Repair Method} & \textbf{Beneficial / Harmful \%} & $\textbf{AUROC}_{\textbf{BCE}} \to \textbf{AUROC}_{\textbf{Cop}}$ & $\overline{\textbf{BIoU}}_{\textbf{orig}} \to \overline{\textbf{BIoU}}_{\textbf{rep}}$ & $\textbf{AP}^{\textbf{mask}}\uparrow$ & $\textbf{AP}^{\textbf{bnd}}\uparrow$ & $\textbf{AP}^{\textbf{box}}\uparrow$ & $\textbf{AP}_{\textbf{S}} / \textbf{AP}_{\textbf{L}}\uparrow$ & $\textbf{ECE}\downarrow$ \\
\midrule
COCO-RealNoise & Original + Standard & -- & 0.811 $\to$ \textbf{0.956} & 70.8\% (88.2\%) & 30.45 & \textbf{17.05} & \textbf{32.03} & 15.61 / 43.21 & 19.12 \\
 & Direct Teacher Replacement + Standard & 50.7\% / 44.8\% & (Frozen 0.629 $\to$ \textbf{0.647}) & 70.8\% $\to$ 68.7\% & 30.31 & 16.61 & 31.82 & 15.94 / 43.73 & 19.70 \\
\rowcolor{bestblue} & \textbf{\texttt{CB-LNL} Cleaned + Standard} & \textbf{51.4\%} / \textbf{38.6\%} &  & 70.8\% $\to$ \textbf{71.5\%} (\textbf{90.1\%}) & \textbf{30.52} & 16.81 & 31.49 & \textbf{16.48} / \textbf{43.83} & \textbf{18.87} \\
\midrule
WAN-Box (Noisy) & Original / \textbf{\texttt{CB-LNL} Cleaned + Std} & \textbf{81.2\%} / \textbf{14.1\%} & 0.767 $\to$ \textbf{0.909} & 70.6\% $\to$ \textbf{76.5\%} & 28.63 $\to$ \textbf{29.27} & \textbf{16.57} / 16.25 & 31.12 $\to$ \textbf{31.20} & \textbf{16.04} / \textbf{41.94} & \textbf{16.10} \\
\midrule
COCO-N Medium & Original + Standard & -- & 0.815 $\to$ \textbf{0.925} & 64.4\% (68.3\%) & 26.70 & 14.34 & 28.46 & 13.39 / 39.11 & \textbf{15.70} \\
 & Direct Teacher Replacement + Standard & 74.6\% / 24.1\% & (Frozen \textbf{0.792} / 0.788) & 64.4\% $\to$ 71.6\% & 27.17 & 14.42 & 27.78 & 13.67 / 40.16 & 16.27 \\
 & \textbf{\texttt{CB-LNL} Cleaned (1-Stage / Multi) + Std} & \textbf{83.3\%} / \textbf{12.8\%} &  & 64.4\% $\to$ \textbf{72.4\%} (\textbf{77.7\%}) & 27.52 / 28.93 & 15.11 / 15.27 & 29.29 / \textbf{30.57} & 14.88 / \textbf{41.72} & 17.79 \\
\rowcolor{bestblue} & \textbf{\texttt{CB-LNL} Cleaned + \texttt{CB-LNL}} &  &  &  & \textbf{29.17} & \textbf{16.17} & 30.56 & \textbf{15.67} / 41.44 & 16.88 \\
\midrule
\textbf{Full-Scale Split} & \textbf{Method (High-Res Schedule)} & $\textbf{AP}_{50}^{\textbf{mask}}\uparrow$ & $\textbf{AP}_{75}^{\textbf{mask}}\uparrow$ & $\textbf{AUROC}_{\text{Cls}} / \textbf{AUROC}_{\text{Bnd}}\uparrow$ & $\textbf{AP}^{\textbf{mask}}\uparrow$ & $\textbf{AP}^{\textbf{bnd}}\uparrow$ & $\textbf{AP}^{\textbf{box}}\uparrow$ & $\textbf{AP}_{S} / \textbf{AP}_{M} / \textbf{AP}_{L}\uparrow$ & $\hat{\lambda}_{\text{final}}$ \\
\midrule
COCO-N Med (Full) & Standard / \texttt{Boundary-LNL} & 46.04 / 47.32 & 29.59 / 30.36 & 0.950 / 0.749 & 28.11 / 29.04 & 15.87 / 16.84 & 31.39 / 32.38 & \textbf{13.68} / 35.58 / 42.00 & -- \\
\rowcolor{bestblue} & \textbf{\texttt{CB-LNL}} / \textbf{\texttt{CB-LNL-EMA}} & 47.63 / \textbf{48.08} & 31.32 / \textbf{31.56} & \textbf{0.958 / 0.756} & 29.44 / \textbf{29.74} & 16.82 / \textbf{17.25} & 32.41 / \textbf{32.66} & 13.34 / \textbf{36.24} / \textbf{44.48} & 0.48 \\
COCO-N Hard (Full) & Standard / \textbf{\texttt{Boundary-LNL}} & 42.42 / \textbf{43.35} & 24.37 / 24.74 & 0.942 / 0.756 & 24.19 / \textbf{24.73} & 11.10 / \textbf{11.87} & 28.66 / 28.84 & 8.85 / \textbf{29.56} / \textbf{40.25} & -- \\
\rowcolor{bestblue} & \textbf{\texttt{CB-LNL}} / \textbf{\texttt{CB-LNL-EMA}} & 42.85 / 42.65 & 24.23 / \textbf{24.76} & \textbf{0.946 / 0.768} & 24.37 / 24.61 & 11.62 / 11.56 & 28.82 / \textbf{28.94} & \textbf{8.93} / 29.04 / 40.15 & 0.57 \\
\bottomrule
\end{tabular}%
}
\end{table*}

Our suite covers three annotation sources. The splits and reference annotations are held fixed within each comparison; \cref{tab:audit_fidelity} summarizes their annotation quality.

\noindent\textbf{COCO-RealNoise: human annotations.}
We pair original COCO annotations with a reference derived from LVIS for the same images, covering $79$ of the $80$ COCO categories. Because LVIS does not verify every category in every image, the training reference retains original COCO annotations where a verified LVIS mask is unavailable. These retained masks are not treated as independent evidence of annotation quality: human contour audits and cleaning-quality measurements use only LVIS-verified pairs. Downstream models are evaluated against the same fixed reference validation set, with the original COCO annotations providing an additional evaluation comparison. \Cref{sec:appendix_audit} details the two reference populations.

\noindent\textbf{COCO-WAN: weak auto-annotations.}
On the same images, we generate SAM2 masks using four prompts: tight boxes, perturbed boxes, center points, and random points. The perturbed-box setting also includes $8\%$ within-supercategory label flips, and the random-point setting includes $10\%$ instance omission. These added errors are part of the benchmark design, rather than errors produced by SAM2 alone. Box perturbations use $\omega\sim\mathcal{U}(1.05,1.25)$.

\begin{figure*}[h!]
    \centering
    \includegraphics[width=0.75\textwidth]{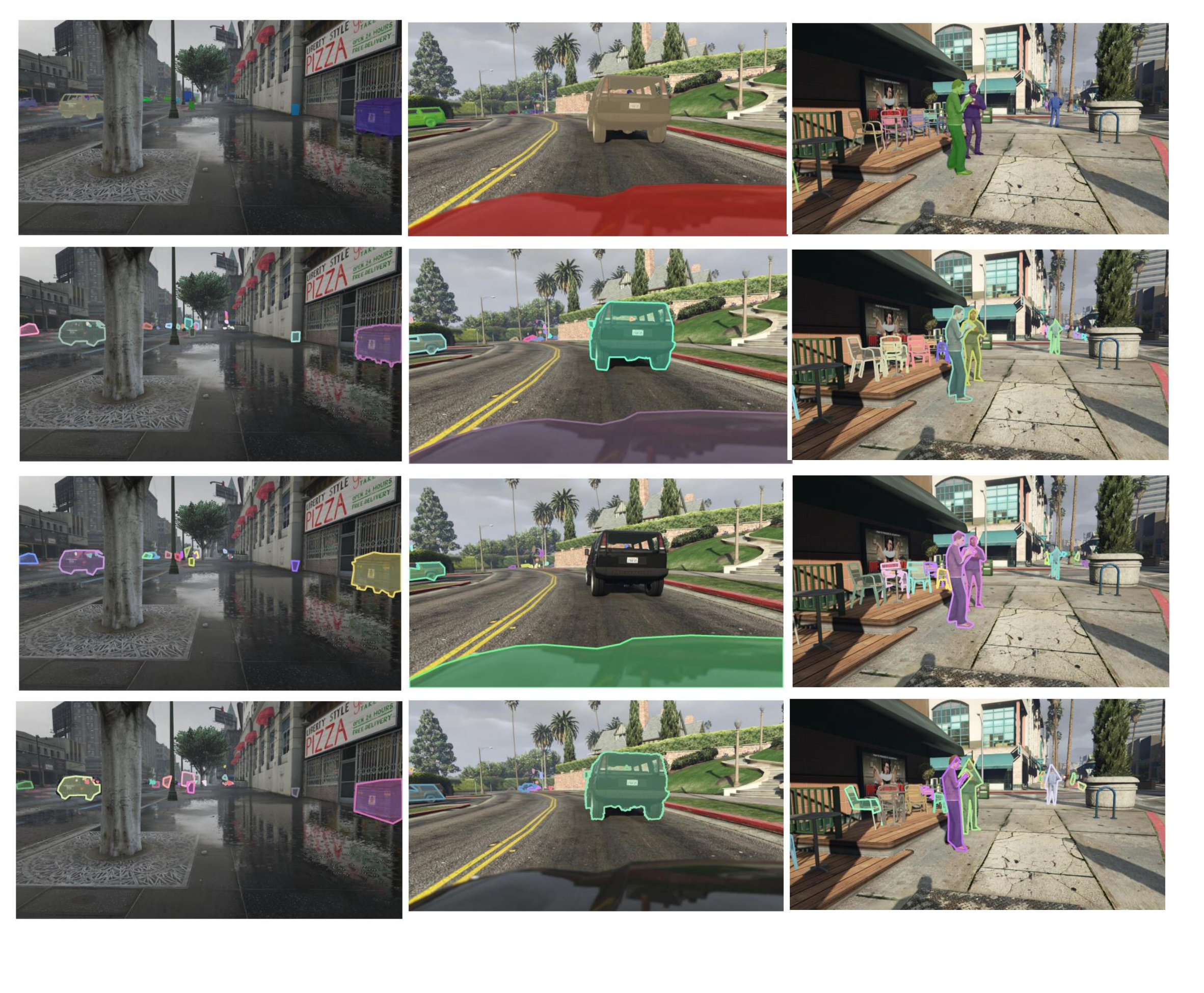}
    \caption{\textbf{Driving-scene examples from VIPER-N.} Reference annotations and procedural corruptions at $1920\times1080$ resolution illustrate displaced contours, simplified polygons, and category errors on vehicles and pedestrians.}
    \label{fig:supp_viper}
\end{figure*}

\noindent\textbf{COCO-N and VIPER-N: controlled corruptions.}
We corrupt annotations in natural scenes from COCO and $1920\times1080$ driving scenes from VIPER~\cite{Richter_2017}. Seven operators test polygon simplification~\cite{douglas1973algorithms}, spatial displacement, vertex jitter, erosion or dilation, within-supercategory label flips, missing instances, and merged neighbors. Their design is guided by the audit. We provide Easy, Medium, and Hard mixtures, ten single-operator splits, and three scale-normalized splits. In the latter, perturbation size scales with the object-box diagonal, $D_i=\sqrt{w_i^2+h_i^2}$. VIPER visual example in \cref{fig:supp_viper}.

\section{Boundary-Aware Learning with Noisy Labels}
\label{sec:method}

\subsection{Why Whole-Mask Loss Misses Contour Errors}
\label{sec:entanglement}
Loss-based sample selection assumes that clean and noisy annotations produce distinguishable losses~\cite{li2020dividemix,arazo2019unsupervised,liu2020early}. For instance segmentation, category errors and contour errors affect different parts of the supervision. Our training-loss diagnostic illustrates the difference (\cref{fig:loss_entanglement}).

\noindent\textbf{Category loss detects category errors, but does not assess the entire annotation.}
On COCO-N Medium, mean classification loss is $5.94$ nats for category-corrupted instances and $1.07$ for clean instances, with an AUROC of $0.95$. A high category loss can therefore be useful evidence of a category error. It does not follow that all mask supervision for that object should be discarded: the mask may still contain useful spatial information.

\noindent\textbf{Whole-mask loss can dilute a local contour error.}
For an $M\times M$ mask grid, let $\mathcal{B}_d(G_i)$ be the width-$d$ band around the annotated contour and let $\mathcal{I}_d(G_i)$ be its complement. The complement includes both foreground and background pixels away from the contour. Full-mask BCE is the weighted average of the two regional losses:
\begin{equation}
\ell_{\text{mask}}^{\text{full}}(P_i, G_i) = \frac{|\mathcal{I}_d|}{M^2} \ell_{\text{BCE}}(\mathcal{I}_d) + \frac{|\mathcal{B}_d|}{M^2} \ell_{\text{bnd}}(\mathcal{B}_d).
\label{eq:loss_decomp}
\end{equation}
A small band contributes little to this average, even when its contour is wrong. In our diagnostic, mean full-mask loss is $0.38$ nats for contour-corrupted masks and $0.26$ for clean masks; detection AUROC ranges from $0.53$ to $0.60$ across settings.

\noindent\textbf{Boundary loss is more informative, but valid edges can also be uncertain.}
Restricting the loss to the boundary band increases detection AUROC to $0.72$--$0.76$ in this diagnostic. The distributions still overlap: mean boundary loss is $0.72$ nats for clean masks and $0.93$--$1.03$ for noisy masks. A prediction near $0.5$ incurs BCE near $\ln 2$ for either binary label, but this is not a universal lower bound on clean-edge loss. We therefore use predictions on both sides of the contour to distinguish an uncertain valid edge from a misplaced annotation.

\subsection{\emph{Sfumato} Supervision}
\label{sec:boundary_lnl}
CB-LNL separates category reliability from contour reliability, then refines supervision near the annotated contour. Its \textbf{Tri-Coupled Fr\'echet Copula} combines three cues: foreground support inside the mask, background support outside it, and agreement between category confidence and mask geometry. These cues use the same pairwise dependence model. \texttt{Boundary-LNL} is the baseline without copula refinement. \texttt{CB-LNL-EMA} adds a teacher whose weights are an exponential moving average of the student weights.

\noindent\textbf{Estimate category and contour reliability separately.}
For each matched foreground region of interest (RoI), we use a $28\times28$ mask grid and a one-pixel boundary band,
$\mathcal{B}_d(G_i)=\mathbb{I}[\operatorname{Dilate}_d(G_i)\ne\operatorname{Erode}_d(G_i)]$, with $d=1$.
Two separate two-component Gaussian mixture models estimate
$\gamma_{\text{cls},i}=\Pr(\text{clean class}\mid\ell_{\text{cls},i})$ and
$\gamma_{\text{bnd},i}=\Pr(\text{clean boundary}\mid\ell_{\text{bnd},i})$ from a rolling loss buffer.
Refinement is restricted to the band; targets elsewhere are left unchanged. This is a local correction rule, not an assumption that every pixel outside the band is correct.

\noindent\textbf{Compare predictions across the contour.}
Averaging all neighbors can mix foreground and background evidence. Instead, we divide the local neighborhood into pixels facing the mask interior, $\mathcal{N}_{\text{in}}(u)$, and pixels facing the exterior, $\mathcal{N}_{\text{out}}(u)$. Let $p_{\text{in}}(u)$ and $p_{\text{out}}(u)$ be their mean predicted foreground probabilities. Then $1-p_{\text{out}}(u)$ is the background evidence outside the mask. We combine pairs of probabilities using
\begin{equation}
C_{\lambda}^{\text{Fr\'echet}}(a, b) = (1 - \lambda)\, a b + \lambda\, \min(a, b), \quad \lambda \in [0, 1].
\label{eq:frechet_copula}
\end{equation}
This copula interpolates between independence ($ab$) and the upper dependence bound ($\min(a,b)$). Nearby predictions share visual features and can be dependent, so treating them as independent can overstate the extra confidence provided by context. For the underlying union term,
\[
\max(a,b)\le a+b-C_\lambda(a,b)\le a+b-ab.
\]
At $\lambda=0$, evidence is combined under independence; at $\lambda=1$, the union equals $\max(a,b)$, so redundant evidence adds no confidence. This bounded, adjustable combination motivates the copula here: it lets context support an uncertain contour without automatically treating neighboring predictions as independent observations. The same function expresses three useful patterns:
\begin{itemize}
    \item \textbf{A supported edge:} high $p_{\text{in}}$ and high $1-p_{\text{out}}$ support the annotated contour, even if the prediction at the contour itself is uncertain.
    \item \textbf{An inward cut:} high foreground probabilities on both sides support extending the mask, measured by $C_\lambda(p_{\text{in}},p_{\text{out}})$.
    \item \textbf{An outward spill:} high background probabilities on both sides support retracting the mask, measured by $C_\lambda(1-p_{\text{in}},1-p_{\text{out}})$.
\end{itemize}
These patterns provide local evidence rather than guarantees: they depend on the model's predictions being informative.

\noindent\textbf{Refine confidence using directional context.}
Following the contextual union formulation used in S4MC~\cite{kimhi2023semisupervised}, we combine a pixel's prediction with directional support:
\begin{equation}
U_{\lambda}^{c}(u) = p^c(u) + \beta_{\text{ctx}}\, g^c(u)\Big[ b^c(u) - C_{\lambda}^{\text{Fr\'echet}}\big(p^c(u), b^c(u)\big) \Big],
\label{eq:copula_union}
\end{equation}
where $c\in\{\text{fg},\text{bg}\}$, $b^c(u)$ denotes directional context support, $g^c(u)$ gates its contribution, and $\beta_{\text{ctx}}$ controls its strength.
We estimate the dependence parameter online as
$\hat\lambda=\operatorname{clip}((J_{\text{emp}}-J_\Pi)/(J_M-J_\Pi+\epsilon),0,1)$.
Here $J_{\text{emp}}$ is the empirical tangential co-occurrence and $J_\Pi$ and $J_M$ are its independence and upper-bound counterparts. This adapts the combination to the observed dependence between predictions.

\noindent\textbf{Account for category uncertainty.}
A class-specific mask head normally supervises channel $Z_i[y_i]$ using the annotated category $y_i$. When that category is unreliable, we also consider the model's candidate category, $\hat c_i=\arg\max_c P_{\text{cls}}(c\mid x_i)$. We score each category's spatial coherence as
$s_{\text{sp}}(c)=\alpha_{\text{in}}\overline P_i^{(c)}|_{\mathcal I_{\text{in}}}+(1-\alpha_{\text{in}})(1-\overline P_i^{(c)}|_{\mathcal I_{\text{out}}})$,
which rewards foreground inside and background outside. We combine it with category confidence using
$J_{\text{cs}}(c)=C_{\hat\lambda}^{\text{Fr\'echet}}(P_{\text{cls}}(c\mid x_i),s_{\text{sp}}(c))$.
When $\gamma_{\text{cls},i}<0.35$, this score guides supervision between $Z_i[y_i]$ and $Z_i[\hat c_i]$. The candidate is a model prediction, not a verified category label.

The same contour scores can be used after training to flag annotations and propose local repairs. We evaluate this cleaning procedure separately from robust training (\cref{sec:exp_suite5}).

\section{Experiments}
\label{sec:experiments}

\subsection{Setup and Metrics}
\label{sec:setup}
We use a common Mask R-CNN head~\cite{he2017mask} with ResNet-50-FPN~\cite{he2015deep,lin2016feature}, ConvNeXt-Tiny-FPN~\cite{liu2022convnext}, or Swin-Tiny-FPN~\cite{liu2021swin}. Training details are in \cref{sec:appendix_impl}. We report Mask AP, size-specific Mask AP ($\text{AP}_S$, $\text{AP}_M$, $\text{AP}_L$), Boundary AP~\cite{Cheng_2021}, Box AP, and expected calibration error (ECE)~\cite{guo2017calibration}. Mask AP measures instance segmentation quality; Boundary AP emphasizes contour alignment.

We first examine how annotation source and corruption type affect training. We then evaluate three separate uses of CB-LNL: robust training, error detection on frozen predictions, and annotation cleaning. Keeping these comparisons separate avoids attributing a training gain to the scoring rule alone.

\subsection{Effects of Annotation Source}
\label{sec:exp_suite1}
\Cref{tab:suite1_archs} compares annotation sources across backbones. Three patterns are consistent across the evaluated models.

\noindent\textbf{Noisy Validation Illusion.}
On the LVIS-derived reference, models trained on reference annotations outperform models trained on original COCO masks. Against the original COCO validation annotations, the ordering reverses. For ResNet-50, for example, original-annotation training scores $29.83$ Mask AP versus $28.31$ for reference-annotation training, although its reference-evaluated score is lower ($30.45$ versus $30.89$). This shows why agreement with the original annotations is not sufficient to establish better contour quality.

\noindent\textbf{Prompt quality affects downstream accuracy.}
Tight-box SAM2 masks give results close to reference-annotation training. Perturbed boxes and single points give lower accuracy across backbones, consistent with the contour and part-selection errors in \cref{sec:wan_audit}.

\noindent\textbf{Contour metrics reveal larger relative losses.}
From clean supervision to Hard corruption, Boundary AP falls proportionally more than Mask AP on both COCO-N and VIPER-N. The trend holds for the evaluated CNN and Transformer backbones (\cref{fig:benchmark_results}a).

\subsection{Effects of Individual Corruptions}
\label{sec:exp_suite2}
\Cref{tab:suite2_ablations} isolates the corruption operators. Applying erosion or dilation consistently reduces Boundary AP from $17.10$ to $8.59$ or $8.63$, respectively. Randomly choosing between them with equal probability gives a much smaller reduction, to $15.29$. The direction of the perturbation therefore matters, in addition to its size.

The tested spatial corruptions also differ substantially: rigid displacement reduces Mask AP from $30.10$ to $25.14$, whereas vertex jitter gives $29.67$. At the tested $20\%$ category-flip rate, Mask AP remains $28.80$--$28.94$. These comparisons identify damaging settings in our benchmark; they do not establish that contour errors always matter more than category errors. Finally, scale-normalized Hard noise reduces large-object Mask AP from $42.18$ to $30.71$, showing that object-relative perturbations can strongly affect large objects too.

\subsection{Training Results Across Seeds}
\label{sec:exp_suite3}
\Cref{tab:suite3_lnl} first compares the methods across six noisy settings using one seed. Loss-based filtering is not reliably helpful: DivideMix-Cls reduces Mask AP in every setting, while DivideMix-Mask has mixed results. CB-LNL improves Mask AP over standard training on COCO-RealNoise and all three COCO-N tiers, but not on the two COCO-WAN settings. Its Boundary AP gains also depend on the annotation source.

We repeat Standard, Boundary-LNL, and CB-LNL using three paired seeds, with splits, corruptions, and initializations matched within each seed (\cref{tab:suite3_seeds}). On \textbf{COCO-N Medium}, CB-LNL reaches $28.57\pm0.27$ Mask AP, compared with $28.12\pm0.16$ for Boundary-LNL and $27.48\pm0.83$ for Standard (mean $\pm$ standard deviation). The paired gain over Boundary-LNL is $+0.44\pm0.38$ Mask AP and is positive in all three runs. Boundary AP differs little: the paired gain is $+0.06\pm0.20$.

On \textbf{COCO-RealNoise}, Boundary-LNL has the higher mean: $30.85\pm0.34$ versus $30.54\pm0.35$ Mask AP for CB-LNL. Both means exceed Standard's $30.19\pm0.26$. Thus, directional refinement adds a clearer benefit on procedural contour noise than on the tested human annotations. Three seeds support this observed pattern, but do not establish that either method will outperform the other across all training conditions.

\subsection{Error Detection and Component Ablations}
\label{sec:exp_suite4}
\Cref{tab:suite4_copula} (top) compares scoring rules on identical frozen predictions from one checkpoint. This isolates scoring from retraining. The held-out sets contain $2,507$ COCO-N Medium objects and $1,123$ LVIS-verified COCO-RealNoise objects. We report AUROC, area under the precision--recall curve (PR-AUC), and the false-positive rate on valid uncertain edges at $80\%$ true-positive rate.

On \textbf{COCO-N Medium}, copula scoring and boundary BCE have similar PR-AUC ($0.8511$ versus $0.8518$), while copula scoring gives fewer false positives on valid uncertain edges ($23.3\%$ versus $28.5\%$). On \textbf{COCO-RealNoise}, copula scoring has the highest AUROC and PR-AUC among the tested rules ($0.6469$ and $0.6776$), although the gains over whole-mask BCE are small. Its false-positive rate is lower than those of the two BCE scores, but higher than linear contrast ($61.9\%$ versus $55.8\%$). The benefit therefore depends on both the dataset and the metric.

Training ablations address a separate question (\cref{tab:suite4_copula}, bottom). The full configuration gives the highest Mask AP in this comparison ($28.45$), exceeding isotropic neighbors ($27.66$) and hard pseudo-labels ($26.89$). It does not give the highest Boundary AP. \Cref{tab:suite4_full_supp} reports the remaining components and dependence-parameter settings.

\subsection{Annotation Cleaning and Full-Scale Training}
\label{sec:exp_suite5}
\label{sec:exp_suite6}
\noindent\textbf{Local repair versus direct teacher replacement.}
\Cref{tab:suite5_cleaning} (top) compares local CB-LNL repair with replacing selected annotations by thresholded $28\times28$ teacher masks. The single-stage comparison uses the same frozen auditor predictions and a matched $30\%$ selection budget. On COCO-N Medium, local repair reduces harmful edits from $24.1\%$ to $12.8\%$ and gives $+0.35$ Mask AP, $+0.69$ Boundary AP, and $+1.51$ Box AP over teacher replacement when the repaired data train a standard Mask R-CNN.

On COCO-RealNoise, local repair also causes fewer harmful edits ($38.6\%$ versus $44.8\%$) and raises LVIS-verified mean Boundary IoU from $70.8\%$ to $71.5\%$. Downstream gains are mixed, however: Mask AP rises slightly over the original annotations ($30.45$ to $30.52$), while Boundary AP falls ($17.05$ to $16.81$). Better annotation overlap does not automatically improve every training metric.

\noindent\textbf{Additional cleaning stages and longer training.}
Multi-stage cleaning on COCO-N Medium reaches $77.7\%$ mean Boundary IoU; combining the cleaned annotations with CB-LNL training gives $29.17$ Mask AP and $16.17$ Boundary AP. These results are reported separately from the matched single-stage comparison.

In full-scale training (\cref{tab:suite6_full}, bottom), CB-LNL-EMA reaches $29.74$ Mask AP and $17.25$ Boundary AP on COCO-N Medium. Its Mask AP at IoU $0.75$ increases by $1.97$ points over Standard. Boundary-LNL remains stronger in overall Mask AP and Boundary AP on the Hard split. The momentum teacher therefore helps in some settings, rather than uniformly resolving severe annotation noise.

\section{Conclusion}
\label{sec:conclusion}
Instance segmentation annotations typically agree over most of an object while disagreeing at its contour, where the manual annotating happens de facto. Our human and SAM2 audits motivate benchmarks that evaluate these errors alongside category and instance-level mistakes. Inspired by \emph{sfumato} painting, CB-LNL uses evidence on both sides of an annotated contour to guide local supervision and repair the annotation masks. Controlled experiments show its clearest gains on procedural contour noise, where classification LNL methods failed due to the difficulty to separate noise from clean labels, while human annotations reveal smaller gains and inherent trade-offs. The broader lesson is to assess category reliability, contour reliability, and repair quality separately, using a fixed validation reference. Semi-supervised learning paradigm with few clean annotations and large-scale unlabeled data remain for future work.

\clearpage
\noindent\textbf{Disclosure of Large Language Model Usage.}
Per CVPR policy, we disclose that a large language model (Gemini-Carbon) assistant was used only for manuscript copy-editing, LaTeX table formatting, and boilerplate code refactoring. All scientific hypotheses, mathematical derivations, manual dataset audits, experimental runs, and empirical analyses were designed, executed, and verified by the authors. Worth to mention that we spend many hours observing annotations and annotating protocols before the use of LLM argentic workflows even existed, and we hope this effort reflected in our research.

{
    \small
    \bibliographystyle{ieeenat_fullname}
    \bibliography{main}
}

\clearpage
\appendix
\setcounter{dbltopnumber}{2}
\renewcommand{\thetable}{S\arabic{table}}
\renewcommand{\thefigure}{S\arabic{figure}}
\setcounter{table}{0}
\setcounter{figure}{0}

\section*{Supplementary material for Sfumato Supervision}

\section{Implementation and Training Details}
\label{sec:appendix_impl}
We implement the models in PyTorch using a common Mask R-CNN FPN v2 head~\cite{he2017mask,lin2016feature}. Backbones are initialized from ImageNet-pretrained ResNet-50~\cite{he2015deep}, ConvNeXt-Tiny~\cite{liu2022convnext}, or Swin-Tiny~\cite{liu2021swin} weights. We use AdamW with initial learning rate $1.2\times10^{-4}$, weight decay $10^{-4}$, cosine learning-rate decay, batch size $8$, and \texttt{bfloat16} mixed precision. SAM2~\cite{ravi2024sam2} receives the box or point prompts specified for each COCO-WAN setting.

For CB-LNL, the category and boundary Gaussian mixture models use a rolling buffer of $2,048$ foreground RoIs, following a one-epoch warm-up. CB-LNL-EMA uses an exponential moving average teacher with decay $0.996$.

\begin{table*}[t]
\centering
\caption{\textbf{Complete architecture metrics.} Values follow ResNet-50 / ConvNeXt-Tiny / Swin-Tiny order for COCO settings, and ResNet-50 / Swin-Tiny order for VIPER-N. Parentheses for COCO-RealNoise use original validation annotations. Other entries use the fixed reference. The main paper presents Mask AP and Boundary AP separately in \cref{tab:suite1_archs}; this table retains all supplied auxiliary metrics.}
\label{tab:suite1_archs_full}
\resizebox{0.93\textwidth}{!}{%
\begin{tabular}{llccccccc}
\toprule
\textbf{Dataset} & \textbf{Supervision} & \textbf{Backbone} & $\textbf{AP}^{\textbf{mask}}\uparrow$ & $\textbf{AP}^{\textbf{bnd}}\uparrow$ & $\textbf{AP}^{\textbf{box}}\uparrow$ & $\textbf{AP}_{\textbf{S}}\uparrow$ & $\textbf{AP}_{\textbf{L}}\uparrow$ & $\textbf{ECE}\downarrow$ \\
\midrule
COCO-RealNoise & Clean Reference & ResNet-50 / ConvNeXt-T / Swin-T & 30.89 (28.31) / 31.26 (28.76) / \textbf{31.34} (28.71) & 17.56 (16.41) / 17.85 (16.58) / \textbf{17.86} (16.57) & 32.02 / 32.16 / \textbf{32.65} & 16.53 / 17.59 / \textbf{17.61} & 44.33 / 45.28 / \textbf{45.80} & 19.20 / 18.88 / \textbf{17.80} \\
 & Noisy COCO & ResNet-50 / ConvNeXt-T / Swin-T & 30.45 (\textbf{29.83}) / 30.69 (\textbf{30.22}) / \textbf{30.94} (\textbf{30.59}) & 17.05 (\textbf{16.82}) / 17.00 (\textbf{16.90}) / \textbf{17.16} (\textbf{17.07}) & 32.03 / 32.25 / \textbf{32.62} & 15.61 / 16.64 / \textbf{17.19} & 43.21 / 44.37 / \textbf{44.57} & 19.12 / 18.70 / \textbf{17.44} \\
\midrule
COCO-WAN & WAN-Box (Clean) & ResNet-50 / ConvNeXt-T / Swin-T & 30.87 / \textbf{31.71} / 31.69 & 18.21 / 18.51 / \textbf{18.76} & 32.00 / \textbf{32.89} / 32.72 & 16.84 / 16.92 / \textbf{17.16} & 44.62 / \textbf{45.82} / 45.55 & 18.92 / 18.94 / \textbf{17.54} \\
 & WAN-Box (Noisy) & ResNet-50 / ConvNeXt-T / Swin-T & 28.63 / 28.98 / \textbf{29.17} & 16.57 / 16.98 / \textbf{17.05} & 31.12 / 31.42 / \textbf{31.66} & 15.82 / 16.17 / \textbf{16.33} & 41.45 / \textbf{42.73} / 42.63 & 16.10 / 16.05 / \textbf{14.87} \\
 & WAN-Point (Center) & ResNet-50 / ConvNeXt-T / Swin-T & 26.21 / \textbf{26.76} / 26.47 & 15.04 / \textbf{15.56} / 15.32 & 26.64 / \textbf{27.14} / 26.82 & 14.53 / 15.00 / \textbf{15.04} & 37.98 / \textbf{38.94} / 38.49 & 19.22 / 19.15 / \textbf{18.14} \\
 & WAN-Point (Random) & ResNet-50 / ConvNeXt-T / Swin-T & 25.00 / 25.43 / \textbf{25.56} & 14.23 / 14.57 / \textbf{14.79} & 25.63 / 25.99 / \textbf{26.24} & 13.79 / 14.12 / \textbf{14.26} & 36.68 / 37.24 / \textbf{37.54} & 18.88 / 18.79 / \textbf{17.85} \\
\midrule
COCO-N & Easy & ResNet-50 / ConvNeXt-T / Swin-T & 29.09 / \textbf{30.10} / 29.55 & 16.29 / \textbf{16.91} / 16.70 & 30.89 / \textbf{31.59} / 30.92 & 14.99 / \textbf{16.08} / 15.49 & 42.33 / \textbf{43.52} / 42.62 & 16.43 / 15.22 / \textbf{14.78} \\
 & Medium & ResNet-50 / ConvNeXt-T / Swin-T & 26.70 / 26.57 / \textbf{27.51} & 14.34 / 14.25 / \textbf{14.97} & 28.46 / 28.18 / \textbf{29.21} & 13.39 / 12.95 / \textbf{13.97} & 39.11 / 38.77 / \textbf{39.67} & 15.70 / 15.71 / \textbf{14.57} \\
 & Hard & ResNet-50 / ConvNeXt-T / Swin-T & 24.09 / \textbf{24.67} / 24.51 & 10.48 / \textbf{10.97} / 10.75 & 26.86 / \textbf{27.32} / 27.14 & 10.42 / \textbf{10.91} / 10.56 & 36.11 / \textbf{36.89} / 36.73 & 14.88 / 14.75 / \textbf{13.95} \\
\midrule
VIPER-N & Clean / Easy & ResNet-50 / Swin-T & 24.19 / \textbf{24.75} (E: 21.86 / \textbf{22.42}) & 15.93 / \textbf{16.34} (E: 14.12 / \textbf{14.56}) & 29.05 / \textbf{29.68} (E: 26.44 / \textbf{27.10}) & 11.42 / \textbf{11.95} & 35.80 / \textbf{36.45} & 14.92 / \textbf{13.84} \\
 & Medium / Hard & ResNet-50 / Swin-T & 19.65 / \textbf{20.14} (H: 17.56 / \textbf{18.02}) & 11.82 / \textbf{12.24} (H: 9.76 / \textbf{10.12}) & 24.18 / \textbf{24.76} (H: 21.95 / \textbf{22.48}) & 8.95 / \textbf{9.32} & 30.12 / \textbf{30.88} & 15.65 / \textbf{14.52} \\
\bottomrule
\end{tabular}%
}
\end{table*}

\section{Reference Annotations and Audit Populations}
\label{sec:appendix_audit}
\Cref{tab:audit_fidelity} summarizes annotation agreement across the benchmark settings. COCO-based audits use LVIS references~\cite{gupta2019lvis}; VIPER-N uses the uncorrupted driving annotations~\cite{Richter_2017}. For human annotations, contour agreement is evaluated only on verified LVIS pairs.

\noindent\textbf{Training and validation references.}
LVIS verifies only selected categories in each image. Our training reference (\texttt{train\_clean\_lvis\_ref.json}) therefore contains both verified LVIS masks and retained COCO masks for categories without a verified reference. It contains $5,000$ images and $41,914$ instances: $19,741$ verified LVIS instances ($47.1\%$; $17,262$ tagged \texttt{lvis\_exhaustive} and $2,479$ tagged \texttt{lvis\_matched}) and $22,173$ retained instances ($52.9\%$; tagged \texttt{coco\_fallback}). This avoids dropping those objects from training. The validation reference (\texttt{val\_clean\_lvis\_ref.json}) has $2,000$ images and $16,253$ instances and remains fixed across comparisons. Fast multi-seed experiments use its first $350$ images.

\noindent\textbf{Human annotation audit.}
The audit spans $24,809$ shared images. Matching against verified LVIS masks uses $\text{IoU}\ge0.30$ for the $59,840$ object pairs and $\text{IoU}\ge0.50$ for the $57,962$ same-class pairs used in contour analysis. The latter have mean Boundary IoU $0.7302$ and median $0.7615$.

\noindent\textbf{Cleaning-quality measurements.}
The cleaning sample contains $8,090$ active instances. Of these, $3,088$ ($38.2\%$) match verified LVIS masks; their mean Boundary IoU is $0.7088$ at matching IoU $0.30$ and $0.7291$ at matching IoU $0.50$. The other $5,002$ ($61.8\%$) match identical retained COCO masks and therefore have Boundary IoU $1$ before repair. Including these self-matches gives an average of $0.8888$ under that calculation ($0.8825$ with full-image morphological erosion). This higher value is agreement with the mixed reference, not independent evidence of accurate human contours.

Accordingly, \cref{tab:suite5_cleaning,tab:supp_cross_audit} distinguish LVIS-verified Boundary IoU from full-split reference agreement. Downstream AP always uses the fixed validation reference. These populations should not be pooled when interpreting annotation quality.

\begin{table*}[t]
\centering
\caption{\textbf{Annotation agreement across benchmark settings.} COCO-based settings use LVIS reference masks; VIPER-N uses uncorrupted driving annotations. Human contour statistics use same-class, LVIS-verified pairs. $\text{BIoU}_S$ and $\text{BIoU}_L$ denote mean Boundary IoU for small ($<32^2$ pixels) and large ($\ge96^2$ pixels) objects. \Cref{sec:appendix_audit} specifies the matching populations.}
\label{tab:audit_fidelity}
\resizebox{\textwidth}{!}{%
\begin{tabular}{llccccccc}
\toprule
\textbf{Dataset Group} & \textbf{Supervision Regime} & \textbf{Mean IoU} & \textbf{Med.\ IoU} & \textbf{Mean BIoU} & \textbf{Med.\ BIoU} & \textbf{\% $(\text{BIoU}<0.75)$} & \textbf{$\text{BIoU}_S$ / $\text{BIoU}_L$} & \textbf{Class Flip / Omission} \\
\midrule
\textbf{Real Human} & \textbf{COCO-RealNoise} (MS-COCO Crowdsourced) & 0.823 & 0.853 & 0.730 & 0.761 & \textbf{47.0\%} & 0.768 / \textbf{0.632} & 3.1\% (88.9\% super) / 24.6\% \\
\midrule
\textbf{COCO-WAN} & \textbf{WAN-Box (Clean)} (SAM2 + Clean GT Box) & \textbf{0.796} & \textbf{0.843} & \textbf{0.741} & \textbf{0.786} & \textbf{41.0\%} & 0.735 / 0.657 & 0.0\% / 0.0\% \\
\textbf{COCO-WAN} & \textbf{WAN-Box (Noisy)} (SAM2 + Noisy Detector Box) & 0.754 & 0.810 & 0.698 & 0.749 & 50.2\% & 0.691 / 0.600 & 8.0\% (super) / 0.0\% \\
\textbf{COCO-WAN} & \textbf{WAN-Point (Center)} (SAM2 + Center Click) & 0.617 & 0.712 & 0.570 & 0.651 & 62.4\% & 0.581 / \textbf{0.430} & 0.0\% / 0.0\% \\
\textbf{COCO-WAN} & \textbf{WAN-Point (Random)} (SAM2 + Random Click) & 0.585 & 0.682 & 0.546 & 0.623 & 64.1\% & 0.563 / \textbf{0.410} & 0.0\% / 10.0\% \\
\midrule
\textbf{COCO-N} & \textbf{COCO-N Easy} (Calibrated to Human Noise) & 0.839 & -- & 0.808 & -- & -- & Corrupted BIoU: 0.728 & 5.0\% / 5.0\% \\
\textbf{COCO-N} & \textbf{COCO-N Medium} (Calibrated to Weak Auto-Labels) & 0.693 & -- & 0.630 & -- & -- & Corrupted BIoU: 0.571 & 8.0\% / 8.0\% \\
\textbf{COCO-N} & \textbf{COCO-N Hard} (Severe Contour \& Semantic Stress) & 0.545 & -- & 0.447 & -- & -- & Corrupted BIoU: 0.419 & 12.0\% / 12.0\% \\
\midrule
\textbf{VIPER-N} & \textbf{VIPER-N Easy} ($1920\times 1080$ Driving) & 0.712 & -- & 0.708 & -- & -- & Corrupted BIoU: 0.583 & 5.0\% / 5.0\% \\
\textbf{VIPER-N} & \textbf{VIPER-N Medium} ($1920\times 1080$ Driving) & 0.528 & -- & 0.522 & -- & -- & Corrupted BIoU: 0.444 & 8.0\% / 8.0\% \\
\textbf{VIPER-N} & \textbf{VIPER-N Hard} ($1920\times 1080$ Driving) & 0.379 & -- & 0.369 & -- & -- & Corrupted BIoU: 0.334 & 12.0\% / 12.0\% \\
\bottomrule
\end{tabular}%
}
\end{table*}

\section{Complete CB-LNL Ablations}
\label{sec:appendix_ablation}
\Cref{tab:suite4_full_supp} expands the main-paper ablations. Part A compares interior evidence, exterior evidence, their directional combination, and category--spatial agreement. Part B varies $\lambda$ from independence ($0$) to the upper dependence bound ($1$), and compares these fixed choices with online estimation. Part C tests hard pseudo-labels, refinement over the full mask, and a larger neighborhood. The table also reports paired training repeats on COCO-N Medium and COCO-RealNoise.

The pass rate is the fraction of contour pixels whose confidence exceeds $0.85$ before or after refinement. A higher pass rate does not necessarily give higher AP. Online estimation gives the highest Mask AP among the listed dependence-parameter configurations, with $\hat\lambda=0.53$ on COCO-N Medium and $0.47$ on COCO-RealNoise.

\begin{table*}[t]
\centering
\caption{\textbf{CB-LNL components and paired training repeats.} Parts A--C vary context evidence, the dependence parameter $\lambda$, and refinement design. Pass$_{\text{pre}}$ and Pass$_{\text{post}}$ are the percentages of contour pixels above confidence $0.85$ before and after refinement. The final rows report individual Mask AP runs and three-seed mean $\pm$ standard deviation; these values are not confidence pass rates.}
\label{tab:suite4_full_supp}
\resizebox{0.96\textwidth}{!}{%
\begin{tabular}{llcccccccc}
\toprule
\textbf{Study Group} & \textbf{Ablation Variant} & \textbf{Copula $\lambda$} & $\textbf{Pass}_{\textbf{pre}} \to \textbf{Pass}_{\textbf{post}}$ & $\textbf{AP}^{\textbf{mask}}\uparrow$ & $\textbf{AP}^{\textbf{bnd}}\uparrow$ & $\textbf{AP}^{\textbf{box}}\uparrow$ & $\textbf{AP}_{\textbf{S}}\uparrow$ & $\textbf{AP}_{\textbf{L}}\uparrow$ & $\textbf{ECE}\downarrow$ \\
\midrule
Baseline & Standard (CE + BCE, Seed 1) & -- & -- & 26.70 & 14.34 & 28.46 & 13.39 & 39.11 & 15.70 \\
Baseline & \texttt{Boundary-LNL} (No Copula, Seed 1) & -- & -- & 28.30 & \textbf{15.48} & 29.74 & 15.24 & \textbf{41.61} & 15.87 \\
\midrule
Part A: Evidence & Isotropic Band Neighbors ($\mathcal{B}_d$ Only) & $\hat{\lambda}=0.53$ & 23.4\% $\to$ 29.3\% & 27.66 & 14.85 & 29.36 & 14.88 & 40.61 & 15.96 \\
Part A: Evidence & Interior Core Only ($p_{\text{in}}$) & $\hat{\lambda}=0.52$ & 23.6\% $\to$ 31.9\% & 28.10 & 15.44 & 29.79 & \textbf{15.52} & 41.31 & 15.72 \\
Part A: Evidence & Inverse-Exterior Only ($1 - p_{\text{out}}$) & $\hat{\lambda}=0.53$ & 23.5\% $\to$ 31.9\% & 27.46 & 14.60 & 29.27 & 14.32 & 39.33 & 15.95 \\
Part A: Evidence & Directional Spatial ($p_{\text{in}} \oplus (1 - p_{\text{out}})$) & $\hat{\lambda}=0.53$ & 23.4\% $\to$ 30.1\% & 27.69 & 15.28 & 29.69 & 15.44 & 40.12 & 16.03 \\
Part A: Evidence & Class--Spatial Copula Only ($C_{\lambda}(P_{\text{cls}}, s_{\text{sp}})$) & $\hat{\lambda}=0.53$ & 0.0\% $\to$ 0.0\% & 28.21 & 15.39 & \textbf{30.02} & 14.78 & 41.00 & \textbf{15.37} \\
\rowcolor{bestblue}Part A: Full & \textbf{\texttt{CB-LNL} (Full Tri-Coupled, Seed 1)} & $\hat{\lambda}=0.53$ & 23.7\% $\to$ 30.7\% & \textbf{28.45} & 15.33 & 29.87 & 14.64 & 41.22 & 15.97 \\
\midrule
Part B: $\lambda$ Sweep & Fixed $\lambda=0.00$ (Independence~\cite{kimhi2023semisupervised}) & $0.00$ & 24.0\% $\to$ \textbf{43.4\%} & 27.96 & 15.02 & \textbf{30.04} & \textbf{15.75} & 40.35 & 15.61 \\
Part B: $\lambda$ Sweep & Fixed $\lambda=0.25$ (Weak Dependence) & $0.25$ & 23.6\% $\to$ 38.0\% & 27.44 & 14.91 & 29.57 & 14.41 & 39.50 & 15.88 \\
Part B: $\lambda$ Sweep & Fixed $\lambda=0.50$ (Moderate Dependence) & $0.50$ & 23.5\% $\to$ 32.8\% & 27.88 & 15.00 & 29.73 & 14.48 & 40.08 & 15.57 \\
Part B: $\lambda$ Sweep & Fixed $\lambda=0.75$ (Strong Dependence) & $0.75$ & 23.4\% $\to$ 27.9\% & 27.35 & 14.72 & 28.62 & 13.82 & 39.47 & 15.74 \\
Part B: $\lambda$ Sweep & Fixed $\lambda=1.00$ (Upper Bound $\min$) & $1.00$ & 23.3\% $\to$ 24.0\% & 27.71 & 15.14 & 29.96 & 14.37 & 41.15 & 15.68 \\
\midrule
Part C: Design & Hard Pseudo-Labels ($\mathbb{I}[U_{\lambda}>0.85]$) & $\hat{\lambda}=0.80$ & 23.5\% $\to$ 27.3\% & 26.89 & 14.88 & 28.70 & 15.04 & 38.83 & 16.01 \\
Part C: Design & Full-Mask Scope ($28\times 28$ RoI) & $\hat{\lambda}=0.80$ & 53.4\% $\to$ 57.7\% & 28.01 & 15.25 & 29.64 & 14.85 & 38.95 & 16.08 \\
Part C: Design & 2-Hop Window ($5\times 5$ Neighborhood) & $\hat{\lambda}=0.80$ & 23.6\% $\to$ 30.0\% & 27.28 & 14.72 & 29.37 & 14.86 & 39.72 & 15.57 \\
\midrule
COCO-N Med (3 Seeds) & Standard (Seeds 1, 2, 3 $\to$ Mean $\pm$ Std) & -- & $[26.70, 27.39, 28.35]$ & $27.48_{\pm 0.83}$ & $14.60_{\pm 0.28}$ & $29.18_{\pm 0.82}$ & $13.89_{\pm 0.48}$ & $40.34_{\pm 1.15}$ & 15.76 \\
COCO-N Med (3 Seeds) & \texttt{Boundary-LNL} (Seeds 1, 2, 3 $\to$ Mean $\pm$ Std) & -- & $[28.30, 28.01, 28.06]$ & $28.12_{\pm 0.16}$ & $15.31_{\pm 0.17}$ & $29.74_{\pm 0.12}$ & $15.10_{\pm 0.18}$ & $\mathbf{41.40_{\pm 0.25}}$ & 15.82 \\
\rowcolor{bestblue}COCO-N Med (3 Seeds) & \textbf{\texttt{CB-LNL} (Seeds 1, 2, 3 $\to$ Mean $\pm$ Std)} & $\hat{\lambda}=0.53$ & $[28.45, 28.88, 28.37]$ & $\mathbf{28.57_{\pm 0.27}}$ & $\mathbf{15.37_{\pm 0.04}}$ & $\mathbf{30.15_{\pm 0.36}}$ & $\mathbf{15.16_{\pm 0.56}}$ & $41.32_{\pm 0.28}$ & 15.91 \\
\midrule
COCO-RealNoise & \texttt{Boundary-LNL} (Seed 1 / 3-Seed Mean $\pm$ Std) & -- & $[30.54, 30.80, 31.22]$ & 30.54 / $\mathbf{30.85_{\pm 0.34}}$ & 17.19 / $\mathbf{17.44_{\pm 0.22}}$ & 31.76 & 16.42 & 41.92 & 19.19 \\
COCO-RealNoise & Fixed $\lambda=0.00$ (Independence~\cite{kimhi2023semisupervised}) & $0.00$ & 30.5\% $\to$ 48.4\% & 29.77 & 16.87 & 31.25 & 16.57 & 42.01 & 18.86 \\
COCO-RealNoise & Fixed $\lambda=1.00$ (Upper Bound $\min$) & $1.00$ & 30.0\% $\to$ 30.6\% & 30.00 & 16.95 & 31.27 & \textbf{17.44} & \textbf{42.86} & \textbf{18.51} \\
\rowcolor{bestblue}COCO-RealNoise & \textbf{\texttt{CB-LNL} (Seed 1 / 3-Seed Mean $\pm$ Std)} & $\hat{\lambda}=0.47$ & $[30.87, 30.18, 30.57]$ & \textbf{30.87} / $30.54_{\pm 0.35}$ & \textbf{17.22} / $17.25_{\pm 0.08}$ & \textbf{32.27} & 16.32 & 42.72 & 19.01 \\
\bottomrule
\end{tabular}%
}
\end{table*}

\section{Annotation Auditing and Failure Cases}
\label{sec:appendix_cases}
\noindent\textbf{Auditing additional annotation splits.}
\Cref{tab:supp_cross_audit} applies the trained auditor without further training. On the $7,999$ audited COCO validation instances, the reported inward-cut, outward-spill, and category-error rates are $3.26\%$, $5.15\%$, and $0.83\%$. After repairing $1,066$ masks, mean Boundary IoU on LVIS-verified pairs increases from $71.12\%$ to $72.04\%$. The larger full-split value also includes retained COCO references and is reported separately.

On unseen center-point SAM2 annotations, repair increases mean Boundary IoU from $65.62\%$ to $72.09\%$. Error rates also vary by shape: the reported fraction below $0.75$ Boundary IoU is $61\%$--$74\%$ for articulated categories such as people and bicycles, compared with less than $19\%$ for the listed compact categories. These are category-level associations, rather than evidence that shape alone causes an error.

The AUROCs in this additional auditing study and the frozen-prediction comparison in \cref{tab:suite4_copula} belong to separate evaluation protocols. They should not be presented as successive improvements on one common test set.

\begin{table*}[t]
\centering
\caption{\textbf{Auditing additional annotation splits without further training.} Columns report error rates, detection AUROC, Precision@100, and Boundary IoU before and after repair. For COCO human annotations, the main Boundary IoU values use LVIS-verified pairs; parentheses give agreement over the full mixed-reference split. This auditing protocol is separate from the frozen-prediction scoring comparison in \cref{tab:suite4_copula}.}
\label{tab:supp_cross_audit}
\resizebox{\textwidth}{!}{%
\begin{tabular}{llccccccc}
\toprule
\textbf{Evaluated Split} & \textbf{Annotation Source} & \textbf{Audited Inst.} & \textbf{Inward Cut \%} & \textbf{Outward Bleed \%} & \textbf{Class Flip \%} & $\textbf{AUROC}_{\textbf{BCE}} \to \textbf{AUROC}_{\textbf{Cop}}$ & \textbf{Prec@100} & $\overline{\textbf{BIoU}}_{\textbf{orig}} \to \overline{\textbf{BIoU}}_{\textbf{clean}}$ \\
\midrule
\textbf{MS-COCO \texttt{val2017}} & Official Validation Benchmark & 7,999 (3,042 LVIS) & 3.26\% & 5.15\% & 0.83\% & 0.8032 $\to$ \textbf{0.9604} & \textbf{100.0\%} & 71.1\% $\to$ \textbf{72.0\%} (88.6\% $\to$ \textbf{90.5\%}) \\
\textbf{MS-COCO \texttt{train2017}} & Official Crowdsourced Training & 8,090 (3,088 LVIS) & 2.72\% & 5.59\% & 0.80\% & 0.8108 $\to$ \textbf{0.9564} & \textbf{100.0\%} & 70.8\% $\to$ \textbf{71.5\%} (88.3\% $\to$ \textbf{90.1\%}) \\
\textbf{WAN-Box (Noisy)} & SAM2 + Jittered Detector Boxes & 8,903 & 4.93\% & \textbf{22.94\%} & 9.15\% & 0.7674 $\to$ \textbf{0.9087} & \textbf{100.0\%} & 70.59\% $\to$ \textbf{76.46\%} (3,296 fixed) \\
\textbf{WAN-Point (Center)} & SAM2 + Single Center Click (Unseen) & 8,903 & \textbf{20.59\%} & 13.47\% & 3.74\% & 0.8398 $\to$ \textbf{0.9467} & \textbf{100.0\%} & 65.62\% $\to$ \textbf{72.09\%} (3,860 fixed) \\
\textbf{COCO-N Medium} & Multi-Factor Procedural Stress & 8,167 & 18.08\% & 16.59\% & 7.48\% & 0.8146 $\to$ \textbf{0.9251} & \textbf{100.0\%} & 64.4\% $\to$ \textbf{72.4\%} (Multi: \textbf{77.7\%}) \\
\bottomrule
\end{tabular}%
}
\end{table*}

\begin{table*}[t]
\centering
\caption{\textbf{Repair performance by error type.} Local contour errors are compared with large missing parts, merged same-category instances, and thin structures. Detection AUROC and repair rates describe the evaluated cases. The gain column reports Boundary IoU changes for contour errors and the separately marked category-AP change for category errors. Examples appear in \cref{fig:supp_cases}.}
\label{tab:supp_failure_taxonomy}
\resizebox{\textwidth}{!}{%
\begin{tabular}{llccccc}
\toprule
\textbf{Regime Type} & \textbf{Error Mode \& Geometric Condition} & \textbf{Governing Copula Mechanism} & \textbf{Detection AUROC} & \textbf{Repair Success Rate} & $\Delta\overline{\textbf{BIoU}}$ \textbf{Gain} & \textbf{Primary Bottleneck / Behaviour} \\
\midrule
\textbf{Positive 1} & Inward Polygon Shortcut (Concavities / Limbs) & Foreground Continuation $C_{\lambda}(p_{\text{in}}, p_{\text{out}})$ & \textbf{0.962} & \textbf{91.8\%} & $\mathbf{+14.2\%}$ & Both sides of cut lie in foreground ($p_{\text{in}}, p_{\text{out}} > 0.8$) \\
\textbf{Positive 2} & Outward Box Bleed (Loose SAM2 / Dilated Contour) & Inverse-Background $C_{\lambda}(1-p_{\text{in}}, 1-p_{\text{out}})$ & \textbf{0.941} & \textbf{88.5\%} & $\mathbf{+11.8\%}$ & Both sides of bleed lie in background ($q_{\text{in}}, q_{\text{out}} > 0.8$) \\
\textbf{Positive 3} & Within-Supercategory Semantic Flip ($\text{IoU}\ge 0.7$) & Class--Spatial Copula $C_{\lambda}(P_{\text{cls}}, s_{\text{sp}})$ & \textbf{0.985} & \textbf{91.4\%} & $\mathbf{+8.9\%}\text{ AP}^{\text{cls}}$ & True class $\hat{c}_i$ verified by inner mask core $\mathcal{I}_{\text{in}}$ \\
\midrule
\textbf{Failure Mode 1} & Extreme Part Truncation ($>40\%$ Missing, \texttt{WAN-Point}) & True contour lies far outside band $\mathcal{B}_d$ ($>6\text{ px}$) & 0.812 & 24.6\% & $+1.9\%$ & Local $r \in \{1,2\}$ band cannot bridge large missing parts \\
\textbf{Failure Mode 2} & Crowded Same-Class Merge (Overlapping Instances) & Shared semantic class on occlusion border & 0.618 & 14.2\% & $+0.6\%$ & Class-specific mask head predicts $p_{\text{in}}, p_{\text{out}} \approx 0.85$ on both \\
\textbf{Failure Mode 3} & Sub-RoI Thin Structures ($<3\text{ px}$ in $28\times 28$ RoI) & Eroded inner core vanishes ($\mathcal{I}_{\text{in}} = \emptyset$) & 0.684 & 19.5\% & $+0.8\%$ & Quantization removes inner anchor $p_{\text{in}}$; defaults to no-op \\
\bottomrule
\end{tabular}%
}
\end{table*}

\begin{figure*}[t]
    \centering
    \includegraphics[width=\textwidth]{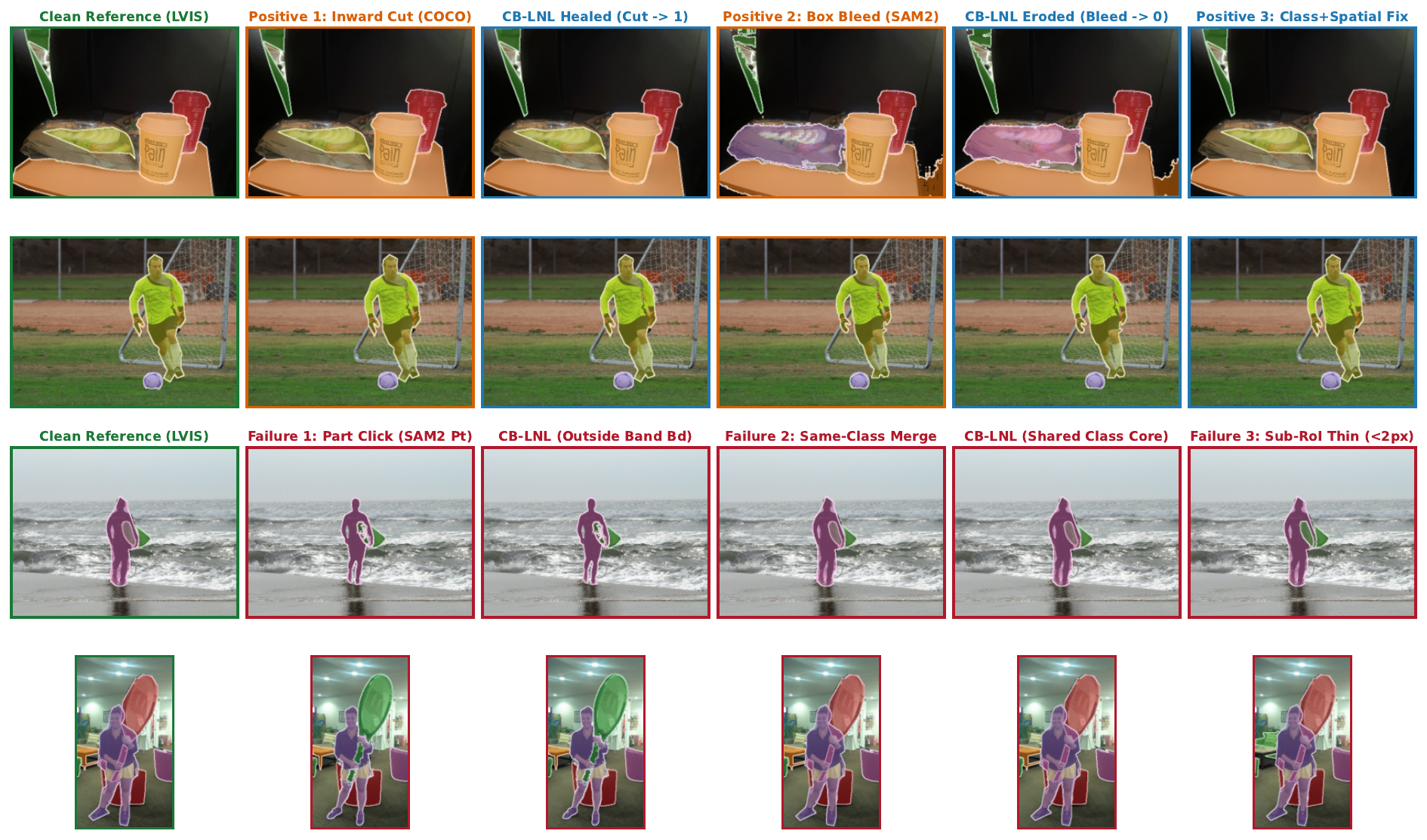}
    \caption{\textbf{Successful repairs and difficult cases.} The top two rows show inward cuts (columns 2--3), outward spills (columns 4--5), and category errors (column 6). The bottom two rows show large missing parts, merged same-category instances, and thin structures. These cases illustrate the limits of relying on local contour evidence (\cref{tab:supp_failure_taxonomy}).}
    \label{fig:supp_cases}
\end{figure*}

\noindent\textbf{Where local repair helps and where it struggles.}
\Cref{tab:supp_failure_taxonomy,fig:supp_cases} compare successful repairs with three difficult cases. In the evaluated examples, inward polygon cuts and outward spills have repair rates of $91.8\%$ and $88.5\%$, respectively. These are local errors for which predictions on both sides of the contour provide useful evidence. Three limitations remain:
\begin{itemize}
    \item \textbf{Large missing parts.} A point-prompt mask may cover only a shirt or windshield. The true outline then lies far beyond the local correction band, so contour refinement cannot recover the full object.
    \item \textbf{Merged instances of the same category.} Two people or cars can produce foreground predictions on both sides of their shared border. Category agreement alone does not identify which pixels belong to which instance.
    \item \textbf{Very thin structures.} At $28\times28$ mask resolution, erosion can remove the interior of a thin pole, spoke, or strap. Without reliable interior evidence, the directional test becomes less informative.
\end{itemize}
These are limitations of the evaluated local method, rather than universal limits of copulas or instance segmentation.

\begin{figure*}[t]
    \centering
    \includegraphics[width=0.96\textwidth]{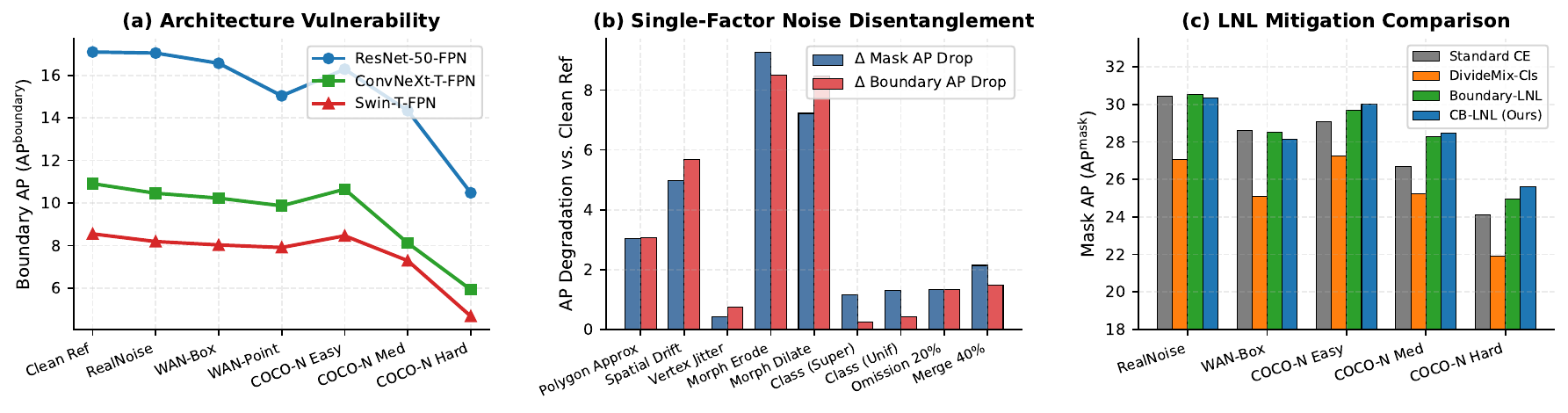}
    \caption{\textbf{Summary of benchmark results.} \textbf{(a)} Boundary AP across backbones and supervision settings (\cref{tab:suite1_archs}). \textbf{(b)} Changes in Mask AP and Boundary AP for individual corruption operators (\cref{tab:suite2_ablations}). \textbf{(c)} Noisy-label methods across the evaluated settings (\cref{tab:suite3_lnl}, single-seed results).}
    \label{fig:benchmark_results}
\end{figure*}

\begin{figure*}[t]
    \centering
    \includegraphics[width=\textwidth]{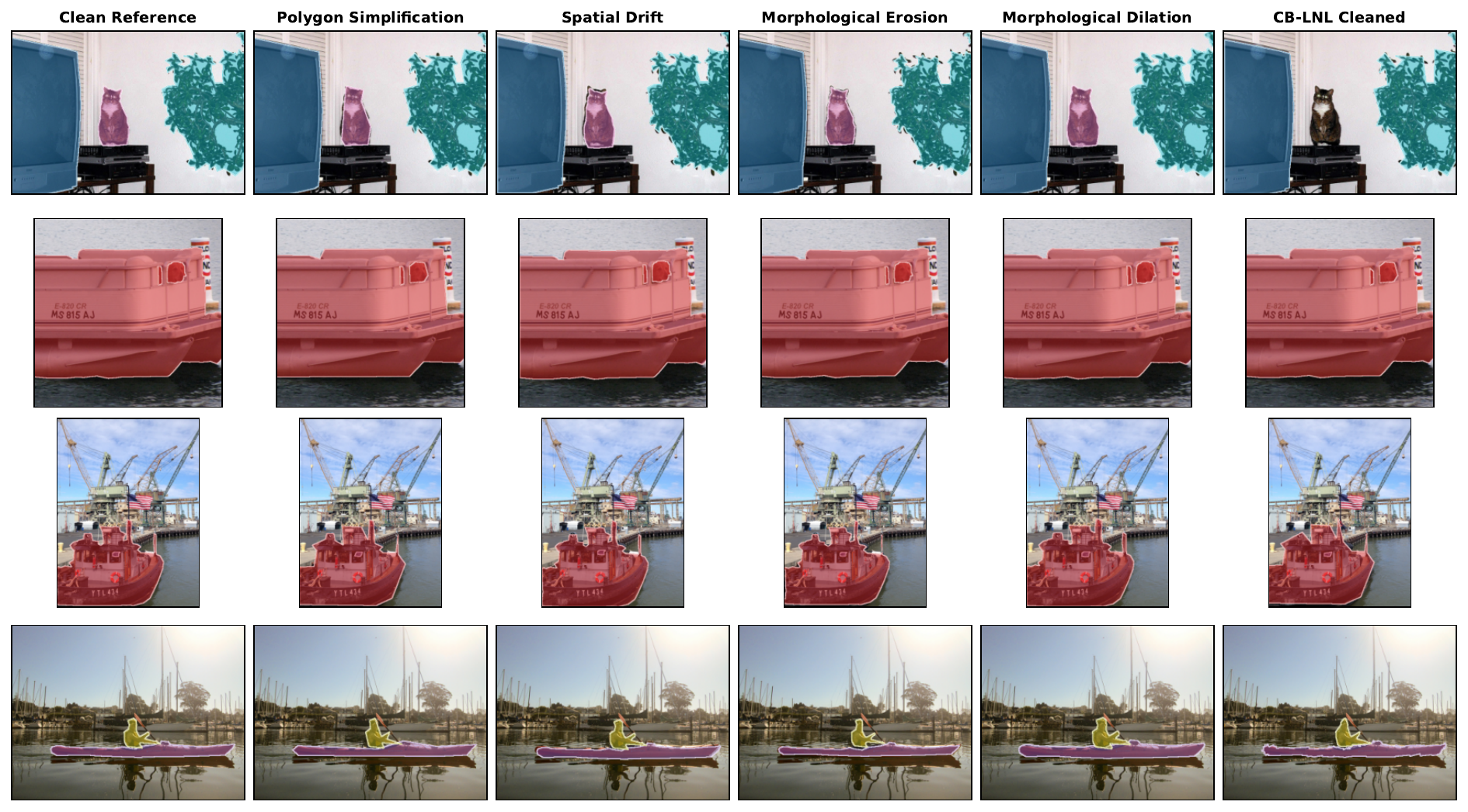}
    \caption{\textbf{Individual contour corruptions and local repair.} Rows show four COCO scenes. Columns compare the reference masks, polygon simplification, spatial displacement, erosion, dilation, and CB-LNL repairs (\cref{tab:suite5_cleaning}).}
    \label{fig:supp_operators}
\end{figure*}

\end{document}